%% file: main.tex
\documentclass[sigconf,nonacm]{acmart}
\usepackage{algorithm}
\usepackage{algorithmic}
\usepackage{array}
\newcolumntype{C}[1]{>{\centering\arraybackslash}p{#1}}

\newcolumntype{L}[1]{>{\raggedright\arraybackslash}p{#1}}
\setcopyright{none}

\title[EvoCast]{EvoCast: Reliable Autonomous Research Agents for Iterative Forecasting Architecture Evolution}
\newcommand{\EvoCastAuthors}{\texorpdfstring{%
Kaipeng Xu\textsuperscript{1}, Xianli Yan\textsuperscript{1},
Yan Wang\textsuperscript{1,*}, Xiang Liu\textsuperscript{1}, Shan Liu\textsuperscript{2}%
}{Kaipeng Xu; Xianli Yan; Yan Wang; Xiang Liu; Shan Liu}}
\author{\EvoCastAuthors}
\affiliation[obeypunctuation=true]{%
\institution{\textsuperscript{1}School of Automation and Intelligent Science, Jiangnan University, Wuxi 214122, China\\
\textsuperscript{2}School of Automation, Southeast University, Nanjing 210096, China}
\city{}
\country{}
\\[3pt]
{\small
\href{mailto:1050324217@stu.jiangnan.edu.cn}{1050324217@stu.jiangnan.edu.cn},
\href{mailto:1051230130@stu.jiangnan.edu.cn}{1051230130@stu.jiangnan.edu.cn}\\
\href{mailto:wangyan88@jiangnan.edu.cn}{wangyan88@jiangnan.edu.cn},
\href{mailto:liux22@jiangnan.edu.cn}{liux22@jiangnan.edu.cn},
\href{mailto:liushan22@seu.edu.cn}{liushan22@seu.edu.cn}\\[2pt]
\textsuperscript{*}Corresponding author: Yan Wang.}
}
\renewcommand{\shortauthors}{Xu et al.}
\hypersetup{pdfauthor={Kaipeng Xu, Xianli Yan, Yan Wang, Xiang Liu, Shan Liu}}

\begin{abstract}
Deep time-series forecasting models have rapidly diversified, yet adapting them to a specific task still requires extensive expert effort in model selection, mechanism diagnosis, architecture design, implementation, and evaluation. Existing AutoML methods are constrained by predefined search spaces, while general-purpose LLM research agents lack reliable control over experimental protocols and model promotion. We introduce EvoCast, a fully autonomous research-agent system for iterative forecasting architecture evolution. EvoCast first establishes and diagnoses a task-specific baseline through executed mechanism ablations, then generates evidence-grounded research directions from dataset characteristics, diagnostic results, prior rounds, and failure records. Its central design, cognition-authority separation, assigns open-ended hypothesis generation and code implementation to LLM agents, while deterministic program authorities control source-edit boundaries, canonical evaluation, and promotion decisions. Experimental outcomes are accumulated as evidence to guide subsequent rounds. Results show that EvoCast completes complex architecture modifications with higher implementation success and lower agent-side token/time cost, and develops task-specific architectures that outperform selected baselines, strong forecasting models, and agent baselines in three real-world forecasting cases.

\par
\noindent The code is available at
\url{https://github.com/18e0-x/EvoCast}.
\end{abstract}


\begin{document}

\maketitle
\hypersetup{pdfauthor={Kaipeng Xu, Xianli Yan, Yan Wang, Xiang Liu, Shan Liu},
pdfsubject={EvoCast arXiv preprint}}

\begin{figure}[t]
\centering
\includegraphics[width=\columnwidth]{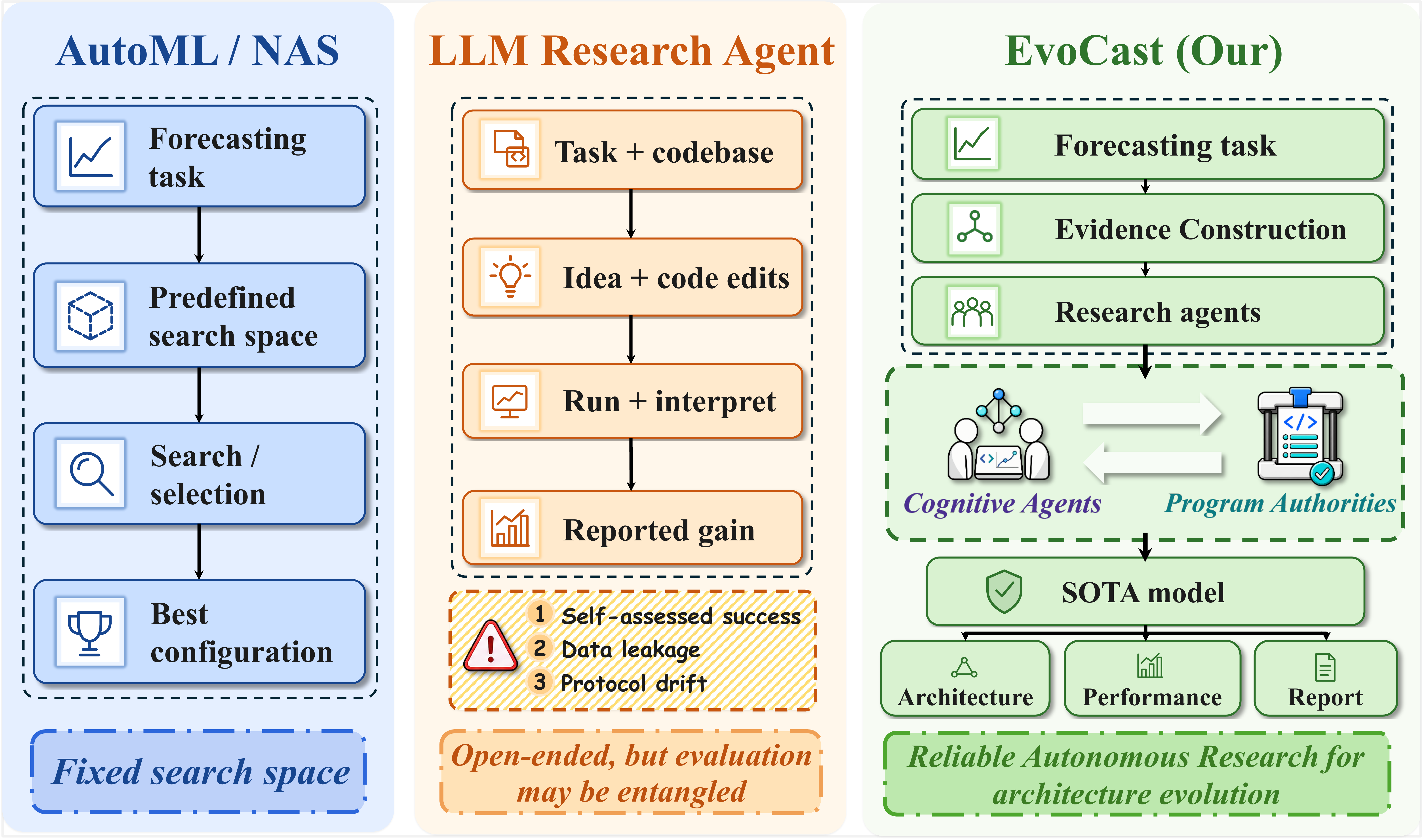}
\caption{EvoCast in the landscape of automated forecasting model development. Compared with fixed-space AutoML/NAS and open-ended LLM research agents, EvoCast combines evidence construction, bounded cognitive-agent exploration, and program-authority verification in a fixed-protocol loop for reliable architecture evolution.}
\Description{A three-part comparison of fixed-space AutoML and neural architecture search, open-ended LLM research agents, and EvoCast. EvoCast links task evidence, bounded cognitive-agent exploration, program-authority verification, canonical evaluation, and evidence-state updates in a closed loop.}
\label{fig:teaser}
\end{figure}

\section{Introduction}
\label{sec:introduction}

Deep time-series forecasting supports decision making in energy systems, transportation systems, industrial monitoring, and many other dynamic environments \citep{lim2021forecasting}. Recent forecasting architectures have rapidly diversified across decomposition-based modeling, frequency-domain modeling, patch tokenization, and variable-centric modeling, forming a heterogeneous and fast-evolving model ecosystem \citep{wu2021autoformer,zhou2022fedformer,nie2023patchtst,liu2024itransformer}. Real forecasting tasks also differ substantially in periodicity, distribution drift, variable interaction, horizon length, and noise characteristics, so a single general-purpose architecture rarely dominates across all settings \citep{qiu2024tfb}. Developing a suitable forecasting architecture for a specific task therefore still requires extensive expert effort, including baseline selection, mechanism diagnosis, architecture modification, source-code integration, and controlled evaluation. A practical autonomous research system must perform these steps under a fixed evaluation protocol so that baselines, candidates, and research rounds remain comparable.

Existing automation methods cover model search, configuration, code refinement, and experiment feedback. Reliable forecasting architecture development additionally requires cross-file information-flow restructuring, task-specific mechanism insertion, fixed-protocol evaluation, and comparable model promotion. These requirements define the setting for EvoCast and distinguish it from existing automated architecture and agentic research systems.

Autonomous forecasting architecture research requires flexible research generation under deterministic experimental control. LLM agents are useful for interpreting heterogeneous forecasting code, identifying candidate mechanisms, and implementing nontrivial architectural edits \citep{huang2024mlagentbench}. Program-controlled experimental rules must fix the dataset, split, metric, training policy, editable source boundary, and promotion rule throughout a run. This requirement builds on the need for fair, consistent, and comparable evaluation emphasized by forecasting benchmarks and NAS evaluation practice \citep{qiu2024tfb,lindauer2020bestnas}. When research generation, code execution, and result adjudication are handled inside the same open-ended agent, failure interpretation, evaluation assumptions, and promotion decisions can contaminate later rounds and waste token and wall-clock budgets \citep{huang2024mlagentbench,xu2026seats}. Unified evaluation infrastructures such as TFB make auditable architecture evolution possible by fixing task definition, evaluation protocol, and model comparison under a common benchmark surface \citep{qiu2024tfb}.

This paper presents EvoCast, a fully autonomous research-agent system for task-specific forecasting architecture evolution. Starting from a forecasting task specification, EvoCast organizes one autonomous run as a fixed-protocol research loop: it establishes a task-specific baseline, diagnoses that baseline through executed mechanism ablations, generates evidence-grounded research directions from dataset characteristics, diagnostic evidence, prior rounds, and failure memory, and implements, evaluates, and promotes candidate architectures under the same protocol. Its central design is cognition-authority separation: LLM agents handle open-ended hypothesis generation, source implementation, and report narration, while deterministic program authorities control source validity, canonical evaluation, metric comparison, model promotion, and state updates. Successful improvements, scientific negative results, and engineering failures are written into the evidence state, reducing repeated trial-and-error and grounding later rounds in auditable facts.

We evaluate EvoCast through three research questions. RQ1 studies implementation reliability and efficiency, asking whether EvoCast can translate forecasting architecture ideas into executable, semantically correct, and protocol-compliant code with higher success and lower agent-side token/time cost. RQ2 studies end-to-end forecasting utility, asking whether EvoCast can develop task-specific architectures that outperform selected baselines, strong forecasting models, and agent baselines in three real-world forecasting cases. RQ3 studies orchestration and authority design, asking which parts of cognition-authority separation are necessary for reliable autonomous architecture research.

Our main contributions can be summarized as follows:
\begin{itemize}
    \item We formulate autonomous forecasting architecture evolution as a fixed-protocol research loop that spans baseline establishment, mechanism diagnosis, bounded implementation, canonical evaluation, promotion, and evidence update in heterogeneous forecasting repositories.
    \item We introduce cognition-authority separation for reliable autonomous research. This design assigns hypothesis generation and source implementation to LLM agents, and deterministic authorities control source validity, runtime evaluation, model promotion, and state updates.
    \item We build an evidence-grounded execution framework that turns research directions into provenance-tracked source candidates, uses repair and failure memory to reduce ineffective iteration, and accumulates scientific and engineering evidence across rounds.
    \item We evaluate EvoCast on repository-level implementation and end-to-end forecasting research tasks, showing stronger implementation reliability, lower token and time cost, and task-specific architecture improvements under controlled budgets.
\end{itemize}

\section{Related Work}
\label{sec:related-work}

\paragraph{Automated Forecasting Architecture Development.}
Deep forecasting architectures encode different assumptions about temporal structure, variable interaction, and information flow \citep{wu2021autoformer,zhou2022fedformer,nie2023patchtst,liu2024itransformer}. TFB shows that their behavior is strongly dataset dependent under a common evaluation protocol \citep{qiu2024tfb}. AutoML, hyperparameter optimization, and neural architecture search automate model selection and configuration within specified templates or operator spaces \citep{he2021automl,elsken2019nas}. Recent LLM-assisted approaches extend this process through multi-agent architecture design, bounded task-specific search, and generated architectures with runtime repair \citep{yang2024nader,jeong2026agentnas,poikarayil2026genautoml}. EvoCast evolves executable source modules from an established task-specific repository baseline, using executed mechanism diagnosis to select bounded architecture changes under a canonical evaluator.

\paragraph{LLMs and Agentic Systems for Time Series.}
Language models support forecasting, temporal reasoning, and multimodal time-series modeling \citep{jin2024timellm,liu2025timecma,zhong2025timevlm,liu2025tsagent,das2026nexus}. TimeSeriesScientist automates data understanding, planning, validation, and reporting, while SEA-TS and GenAutoML develop forecasting code or task-specific architectures through iterative feedback \citep{zhao2025timeseriesscientist,xu2026seats,poikarayil2026genautoml}. EvoCast combines an executable task baseline, executed baseline ablations, persistent evidence state, and deterministic promotion authority, so each subsequent architecture direction is grounded in formal evidence from the same protocol.

\paragraph{Autonomous Research Agents and Controlled Experimentation.}
Autonomous research agents connect research proposals, code editing, and experimental feedback into iterative workflows. MLAgentBench provides standardized machine-learning experimentation environments, AIDE explores candidates through code-space search, R\&D-Agent coordinates research and development roles, and AutoResearch verifies candidates through execution feedback \citep{huang2024mlagentbench,jiang2025aide,yang2025rdagent,kumar2026autoresearch}. MLE-STAR further uses executed ablations to locate high-impact code regions for targeted refinement \citep{nam2025mlestar}.

For long-horizon research, ResearchLoop, AIRA\textsuperscript{2}, FML-bench, ResearchGym, and Arbor emphasize task contracts, consistent evaluation, execution infrastructure, and cross-round evidence management \citep{xia2026researchloop,hambardzumyan2026aira2,zou2026fmlbench,garikaparthi2026researchgym,jin2026arbor}. EvoCast applies these principles to task-specific forecasting architecture evolution: starting from an executable repository baseline, live-path mechanism diagnosis and cross-round evidence guide bounded source changes, while deterministic programs control boundary validation, canonical evaluation, model promotion, and evidence write-back. This design connects research-direction generation with experimental adjudication through explicit role separation, keeping architecture changes and subsequent evidence accumulation under one fixed protocol. Appendix~\ref{app:related-work} expands the comparison.

\section{Preliminaries}
\label{sec:preliminaries}

\begin{figure*}[t]
\centering
\includegraphics[width=0.85\textwidth]{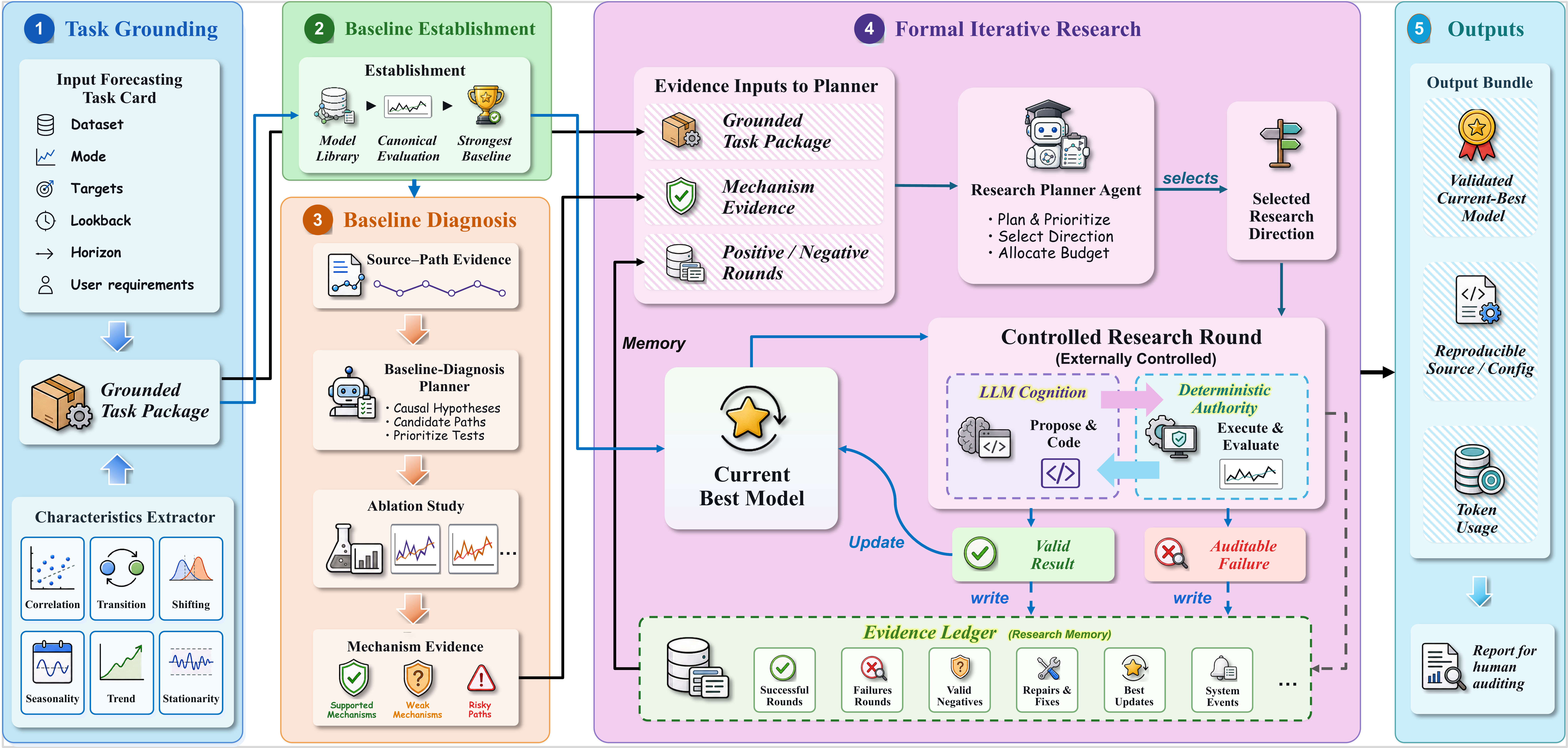}
\caption{Complete EvoCast workflow. The loop fixes the task protocol, establishes and diagnoses a task-specific baseline, executes formal research rounds under bounded implementation and canonical evaluation, updates the evidence-carrying research state, and produces a final HTML report.}
\Description{Workflow diagram with five stages: task grounding, baseline establishment, baseline diagnosis, formal research rounds, and final reporting. Research directions pass through a build contract, an isolated agent workspace, source audit, canonical evaluation, and a promotion gate before their outcomes update the evidence state.}
\label{fig:method_workflow}
\end{figure*}

\subsection{Task Setting}

EvoCast organizes autonomous forecasting architecture development under a fixed task protocol. For one target task, the system first freezes
\[
\mathcal{P}=(\mathcal{D},y,L,H,\mathrm{mode},m,B),
\]
where $\mathcal{D}$ is the dataset, $y$ is the prediction target, $L$ is the lookback length, $H$ is the forecasting horizon, $\mathrm{mode}$ is the forecasting mode, $m(\cdot)$ is the formal evaluation metric, and $B$ is the research budget. This protocol remains unchanged throughout one autonomous run and is reused by baseline selection, diagnosis ablations, formal research candidates, and final promotion checks.

This formulation defines EvoCast as a repository-level research problem under one stable comparison surface. The system must operate on real heterogeneous forecasting repositories, produce executable architecture modifications, and return a validated current-best architecture together with the executed research evidence that supports it.

\subsection{Research State and Round Outcome}

EvoCast organizes one autonomous run as a stateful multi-round research process. At round $r$, the system maintains
\[
\mathcal{S}_r=(M_r,E_r,\mathcal{H}_r),
\]
where $M_r$ is the current incumbent model, $E_r$ is the evidence bundle used for planning, and $\mathcal{H}_r$ is the round history. The evidence bundle aggregates task evidence, baseline-diagnosis evidence, accepted results, valid negatives, unstable gains, repair records, and failure memory. The round history preserves the executed research trajectory for later planning and state update.

Each formal research round returns an outcome
\[
o_r \in \{\mathrm{accepted},\mathrm{rejected},\mathrm{unstable},\mathrm{failed}\}.
\]
An accepted round passes boundary validation, produces valid canonical metrics, and satisfies the fixed multi-seed promotion rule against the incumbent. A rejected round produces valid canonical metrics but does not improve over the incumbent. An unstable round improves in a single run but does not satisfy the seed-stability requirement for promotion. A failed round does not produce a valid formal result because source construction or canonical execution fails. These outcomes give EvoCast a unified state-update interface in which every round contributes auditable evidence.

\section{Method}
\label{sec:method}

\subsection{Overview of the EvoCast Research Loop}

EvoCast organizes task-specific forecasting architecture development as a fixed-protocol autonomous research loop. Given a task card, a forecasting repository, and a bounded research budget, the system follows the five stages in Figure~\ref{fig:method_workflow}: task grounding, baseline establishment, baseline diagnosis, formal iterative research, and final reporting. The loop is driven by the frozen task protocol $\mathcal{P}$ and the evolving research state $\mathcal{S}_r$.

At round $r$, EvoCast reads $\mathcal{S}_r$, generates one research direction, executes one bounded candidate, and writes the validated outcome into $\mathcal{S}_{r+1}$. The equations and algorithm below define this transition.

\subsection{Task Grounding and Baseline Establishment}

Task grounding turns the task card into a stable research starting point. EvoCast reads the dataset, target, lookback, horizon, forecasting mode, and evaluation metric, freezes them as $\mathcal{P}$, and writes them into the grounded task package. The characteristics extractor then computes dataset characterization features and adds them to the task evidence, so later planning is conditioned on executed task structure.

Baseline establishment defines the single formal comparison anchor for the entire run. Let $\mathcal{B}_{\mathrm{exec}}$ denote the set of baseline models that execute successfully under protocol $\mathcal{P}$. EvoCast selects
\[
b^{\star}=\arg\min_{b\in\mathcal{B}_{\mathrm{exec}}} m(b;\mathcal{P}),
\]
where the metric direction follows the formal evaluation setting. In the experiments of this paper, lower error indicates a stronger model. The selected single baseline anchors all later candidate comparisons and also serves as the execution starting point for mechanism diagnosis.
Under this design, the canonical evaluator reuses the same dataset split, metric, training policy, and evaluation path for baselines, ablations, and promoted candidates, thereby fixing one comparison surface throughout the run.

\subsection{Baseline Diagnosis for Mechanism Evidence}

The selected baseline identifies which executable model is currently strongest, and baseline diagnosis identifies why that model works on the current task. EvoCast analyzes the active prediction path of $b^{\star}$, extracts ablatable mechanisms, and turns them into executable diagnosis targets tied to concrete source regions and input-output preservation contracts.

Let $\mathcal{A}(b^{\star})$ denote the mechanism set extracted from the active path of the selected baseline. For each diagnosis target $a\in\mathcal{A}(b^{\star})$, EvoCast executes the ablation under the same protocol and records
\[
\Delta(a)=m(a;\mathcal{P})-m(b^{\star};\mathcal{P}).
\]
Ablation outcomes provide mechanism-level evidence for later planning. Clear degradation after removal supports preserving the mechanism's core function; improvement suggests redundancy or task mismatch and identifies a replacement or restructuring target; a small change marks a weak or unresolved mechanism. Build, runtime, and protocol failures record implementation risks, while mechanism judgments rely on ablations with valid formal metrics. Diagnosis thus identifies functions to preserve, mechanisms to explore, and implementation risks to avoid.

\subsection{Formal Iterative Research with Cognition-Authority Separation}

Formal iterative research is the largest closed-loop region in Figure~\ref{fig:method_workflow}. At the start of round $r$, the Research Planner reads the frozen task protocol and the current research state, then produces one research direction
\[
d_r=\Pi(\mathcal{P},\mathcal{S}_r),
\]
where $\Pi(\cdot)$ denotes the planner policy. The planner uses different evidence sources for different decisions. Task evidence describes periodicity, trends, variable scale, and horizon requirements; the current model and its active source path establish the executable starting point; diagnostic ablations, accepted results, valid negatives, and unstable gains update mechanism judgments; import, shape, runtime, and boundary failures identify implementation risks. The planner first identifies unresolved architectural needs, then generates multiple standalone module-level hypotheses. It compares their evidence grounding, information-flow changes, redundancy with prior directions, and implementation feasibility before selecting one direction for execution.

The selected direction $d_r$ specifies the target mechanism, source insertion point, testable hypothesis, expected information-flow change, failure to avoid, and expected knowledge gain. The hypothesis links task needs to the architecture modification, while expected knowledge gain states which mechanism explanation the outcome would support or weaken and how it would inform later direction selection. Historical rounds supply evidence for this judgment, and implementation starts from the active source state. Each round executes one direction, tying its outcome to a specific research hypothesis.

The builder then compiles $d_r$ into a bounded implementation task. This task fixes the active source snapshot, editable boundary, task protocol, and repair budget. It allows new-module insertion, information-flow rerouting, and local path restructuring inside the allowed source region, while keeping the task semantics, training policy, evaluation interface, and public model contract unchanged.
The builder edits only the allowed model-side source region, while the dataset path, evaluator, metric computation, training budget, and promotion logic remain frozen in the authority-controlled execution path.

\begin{figure}[t]
\centering
\includegraphics[width=\columnwidth]{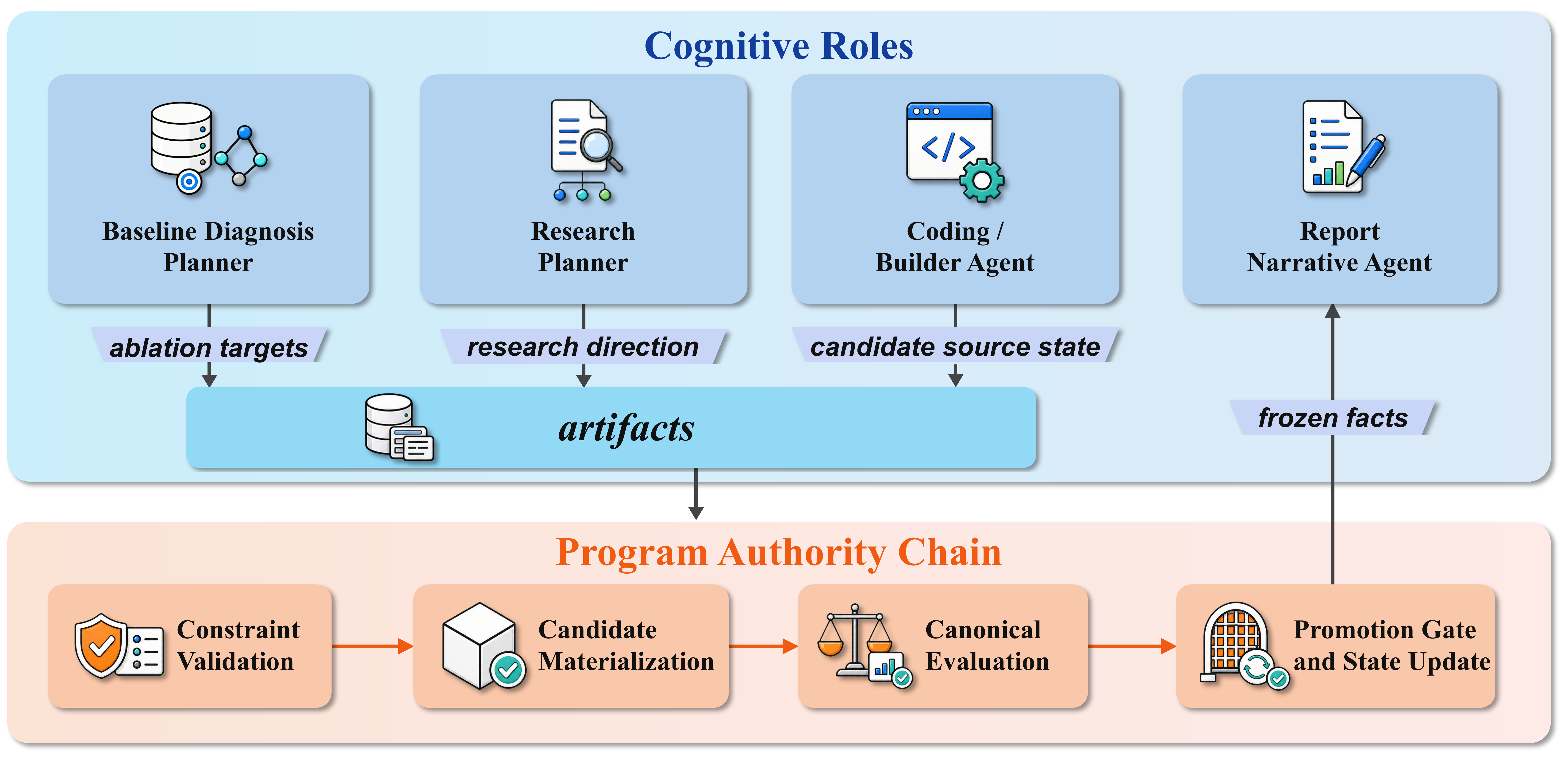}
\caption{Cognitive roles and program authority chain in one controlled research round. Cognitive agents generate diagnosis targets, research directions, candidate source states, and final narrative, while program authorities enforce source boundaries, canonical evaluation, promotion decisions, and state updates.}
\Description{Diagram of a controlled research round. Cognitive agents create research content and source edits. Program authorities audit the source boundary, materialize the candidate, run canonical evaluation, make promotion decisions, and update persistent state.}
\label{fig:method_role_authority}
\end{figure}

Figure~\ref{fig:method_role_authority} shows the cognition-authority boundary inside one controlled research round. Cognitive agents generate research content, and program authorities produce the formal experimental outcome. Algorithm~\ref{alg:main_round} summarizes one evidence-grounded research round.

\begingroup
\setlength{\intextsep}{10pt}
\begin{algorithm}[H]
\caption{One evidence-grounded research round under the fixed EvoCast protocol.}
\label{alg:main_round}
\small
\begin{algorithmic}[1]
\REQUIRE Fixed protocol $\mathcal{P}$, current research state $\mathcal{S}_r$
\ENSURE Updated research state $\mathcal{S}_{r+1}$
\STATE $d_r \leftarrow \Pi(\mathcal{P}, \mathcal{S}_r)$
\STATE $c_r \leftarrow \beta(d_r, \mathcal{P}, M_r)$
\STATE Validate source boundary and materialize canonical candidate $c_r$
\STATE $o_r \leftarrow \Gamma(\mathcal{P}, \mathcal{S}_r, c_r)$
\IF{$o_r = \mathrm{accepted}$}
\STATE Promote $c_r$ to the incumbent model
\ENDIF
\STATE $\mathcal{S}_{r+1} \leftarrow U(\mathcal{S}_r, c_r, o_r)$
\STATE \textbf{return} $\mathcal{S}_{r+1}$
\end{algorithmic}
\end{algorithm}
\endgroup

\subsection{Validated Outputs and Evidence-Carrying State Update}

After direction selection, candidate execution follows the two-step form:
\[
\quad c_r=\beta(d_r,\mathcal{P},M_r),
\qquad
o_r=\Gamma(\mathcal{P},\mathcal{S}_r,c_r),
\]
where $\beta(\cdot)$ denotes bounded candidate construction under the fixed protocol and editable source region, and $\Gamma(\cdot)$ denotes authority-controlled execution. This chain performs source-boundary validation, canonical training and evaluation, metric comparison, and promotion adjudication. Only candidates that pass boundary validation, produce valid formal metrics, and satisfy the fixed comparison rule can enter the accepted state. Model-state promotion is triggered only when the candidate passes boundary validation, produces valid canonical metrics, and satisfies the fixed multi-seed comparison rule against the current incumbent.

The incumbent model is then updated by
\[
M_{r+1}=
\left\{
\begin{array}{ll}
c_r, & \mbox{if } o_r=\mathrm{accepted},\\
M_r, & \mbox{otherwise}.
\end{array}
\right.
\]
and the next research state is written as
\[
\mathcal{S}_{r+1}=U(\mathcal{S}_r,c_r,o_r).
\]
The update $U(\cdot)$ writes model state, mechanism evidence, and engineering memory according to the outcome. An accepted candidate becomes the incumbent, with its supporting formal metrics and gate decision recorded. A rejected candidate leaves the incumbent unchanged and records an executed hypothesis that did not improve the metric as a valid negative. An unstable candidate preserves the incumbent and records both the single-run gain and the failed stability gate. A failed candidate preserves the incumbent and records the failure stage, cause, and repair trace. Valid negatives adjust mechanism judgments, unstable gains indicate evidence strength, and engineering failures identify implementation risks.

The next round rereads these records alongside the current model to assess supported explanations, weakened directions, and unresolved needs. Model state changes only after promotion, while evidence state accumulates valid improvements, negatives, unstable gains, and failure records, allowing subsequent planning to use the full research trajectory.

When the research budget is exhausted or the termination condition is met, EvoCast freezes the final incumbent, formal metrics, accepted rounds, valid negatives, gate decisions, and failure records. The Report Narrative Agent reads only these frozen facts and generates the final HTML report. The output bundle therefore contains the validated current-best model, reproducible source and configuration, resource usage, and a human-auditable final narrative. Appendix~\ref{app:system} details the prompts, execution lifecycle, repair, and state updates.

\section{Experiments}
\label{sec:experiments}

\begin{table*}[t]
\centering
\small
\caption{Overall method-level RQ1 results. Success is the proportion of tasks that pass the canonical validation pipeline; Fidelity is the proportion of evaluated implementations that realize the given idea without semantic mismatch; Repair is computed only over repair-eligible tasks. Higher is better for Success, Fidelity, and Repair; lower is better for Avg. Token, Avg. Cost, and Avg. Time. Best values are \textbf{bolded} and second-best values are \underline{underlined}.}
\label{tab:rq1_main_results}
\begin{tabular}{p{0.25\textwidth}cccccc}
\toprule
Method & Success & Fidelity & Repair & Avg. Token & Avg. Cost (USD) & Avg. Time (s) \\
\midrule
Direct Independent Edit & 33.3\% (10/30) & 90.0\% & --- & \textbf{4,975} & \textbf{0.0037} & \textbf{31} \\
mini-SWE-agent (2024) & \underline{90.0\% (27/30)} & 92.6\% & \underline{90.0\% (27/30)} & 187,818 & 0.0206 & 573 \\
AIDE (2025) & 83.3\% (25/30) & 96.0\% & 54.5\% (6/11) & 38,483 & 0.0356 & 283 \\
R\&D-Agent (2025) & 46.7\% (14/30) & \textbf{100.0\%} & 42.9\% (12/28) & 60,932 & 0.0313 & 266 \\
AutoResearch (2026) & 86.7\% (26/30) & 96.2\% & 50.0\% (4/8) & 38,780 & 0.0327 & 258 \\
EvoCast & \textbf{96.7\% (29/30)} & \textbf{100.0\%} & \textbf{95.7\% (22/23)} & \underline{27,618} & \underline{0.0137} & \underline{153} \\
\bottomrule
\end{tabular}
\end{table*}

\begin{table*}[t]
\centering
\small
\caption{RQ2 task protocols and forecasting characteristics. The Lookback $\rightarrow$ Horizon column gives each task's history and prediction windows. All tasks use chronological 7:1:2 train/validation/test splits.}
\label{tab:rq2_tasks_main}
\begin{tabular*}{\textwidth}{>{\raggedright\arraybackslash}m{0.15\textwidth}@{\extracolsep{\fill}}>{\centering\arraybackslash}m{0.04\textwidth}>{\centering\arraybackslash}m{0.10\textwidth}>{\centering\arraybackslash}m{0.15\textwidth}>{\raggedright\arraybackslash}m{0.44\textwidth}}
\toprule
Task & Mode & Selected backbone & Lookback $\rightarrow$ Horizon & Forecasting characteristics \\
\midrule
Air Quality & MM & FreTS & $168 \rightarrow 24$ & Cross-variable dependency, seasonal structure, and trend variation across 12 air-quality, sensor-response, and weather variables. \\
\addlinespace[4pt]
Melbourne Pedestrian T33 & SS & PatchMLP & $168 \rightarrow 24$ & Hourly pedestrian counts with strong daily and weekly periodicity and long-term level change. \\
\addlinespace[4pt]
Steel Industry & MS & PatchMLP & $720 \rightarrow 120$ & Industrial power forecasting from 9 input variables to Usage\_kWh, with a long lookback and horizon, regime transitions, seasonal structure, and distribution drift. \\
\bottomrule
\end{tabular*}
\end{table*}

\subsection{Research Questions}

We organize the evaluation around three research questions.
\begin{itemize}
    \item \textbf{RQ1: Implementation Reliability.} Can EvoCast faithfully translate forecasting architecture ideas into executable, semantically correct, and protocol-compliant implementations across heterogeneous forecasting repositories?
    \item \textbf{RQ2: End-to-End Research Utility.} Can EvoCast autonomously discover task-specific architectures that improve over strong forecasting and research-agent baselines under controlled budgets?
    \item \textbf{RQ3: Orchestration and Authority Design.} Which parts of cognition-authority separation are necessary for reliable autonomous architecture research?
\end{itemize}

\subsection{Experimental Setup}

All experiments are conducted on the unified benchmark infrastructure provided by TFB. Baseline, ablation, and candidate evaluation reuse the same canonical forecasting benchmark path, and the comparison does not change the evaluator, task definition, or editable boundary. All cognitive-agent calls use MiniMax-M3 as the shared foundation model backend.

Experiments run on Ubuntu 22.04 with Python 3.12.7, PyTorch 2.7.1, CUDA 12.8, and one NVIDIA RTX 3080 Ti GPU (12 GB). API costs use the MiniMax-M3 standard-tier rates applied in our experiments: 0.30, 0.06, and 1.20 USD per million uncached input, cache-hit input, and output tokens, respectively. Appendix~\ref{app:protocols} provides the detailed evaluation protocol, budget-accounting boundaries, and supported baseline pool.

\subsection{RQ1: Implementation Reliability}

RQ1 evaluates whether a method can reliably realize a given forecasting architecture idea as executable code. We construct a controlled idea-to-code benchmark with 30 tasks from 10 forecasting backbones and three predefined idea-complexity levels: local modification, cross-module integration, and novel module insertion. All methods receive the same repository snapshot, idea text, and editable boundary. The compared methods are Direct Independent Edit, mini-SWE-agent \citep{yang2024sweagent}, AIDE \citep{jiang2025aide}, R\&D-Agent \citep{yang2025rdagent}, and AutoResearch \citep{kumar2026autoresearch}. This setup directly compares their ability to implement the same architecture idea in real heterogeneous forecasting repositories.

The three complexity levels operationalize distinct implementation capabilities. Local modification requires a precise edit inside an existing deep-learning module, such as replacing an activation function or inserting a lightweight channel mixer. Cross-module integration requires understanding multiple source regions and establishing a new executable connection, such as adaptive fusion between a decomposition module and a forecasting head. Novel module insertion requires the new mechanism to participate in the live forecasting path or to reorganize that path. This stratification makes success rate reflect local coding precision, cross-file interface matching, and main-path restructuring ability.

Table~\ref{tab:rq1_main_results} reports the overall results. EvoCast achieves the highest success rate while maintaining perfect fidelity and the strongest repair performance, and it is also more time- and cost-efficient than the other agent baselines. Direct Independent Edit is the cheapest method, but it also has the lowest success rate; the remaining agent baselines outperform direct editing but still fall clearly behind EvoCast. These results show that EvoCast improves success rate, efficiency, and stability together when implementing complex deep-learning architecture changes. As tasks move from local edits to cross-module integration and novel module insertion, cognition-authority separation more reliably turns research intent into valid implementations. Appendix~\ref{app:rq1} reports the complexity-level and backbone-level analysis.

\subsection{RQ2: End-to-End Research Utility}

RQ2 evaluates EvoCast on three real forecasting cases: Air Quality, Melbourne Pedestrian T33, and Steel Industry, corresponding to MM, SS, and MS settings. MM denotes multivariate-to-multivariate, SS single-series, and MS multivariate-to-single-target forecasting. Each rolling example uses a history window of length $L$ to forecast the following horizon of length $H$. All three tasks are built on public datasets: Air Quality uses hourly pollutant concentrations and gas-sensor responses from a road-level monitoring station in an Italian city \citep{devito2008airquality}, Melbourne Pedestrian T33 uses public hourly pedestrian counts from the City of Melbourne pedestrian counting system \citep{godahewa2020melbourne}, and Steel Industry uses public electricity-consumption records from a smart small-scale steel industry in South Korea \citep{sathishkumar2021steel}. The compared methods include Sundial \citep{liu2025sundial}, Toto-2.0-2.5B \citep{datadog2026toto2}, Time-VLM \citep{zhong2025timevlm}, TimeCMA \citep{liu2025timecma}, AIDE (2025) \citep{jiang2025aide}, R\&D-Agent \citep{yang2025rdagent}, Selected baseline, and EvoCast, where Selected baseline denotes the strongest executable baseline selected by EvoCast under the fixed forecasting protocol at the start of each task. For agent methods, the system first completes the full workflow and then reevaluates the final selected program under the unified forecasting protocol, so this section compares final deliverable forecasting artifacts under one shared evaluator.

Table~\ref{tab:rq2_tasks_main} lists FreTS, PatchMLP, and PatchMLP as the respective starting backbones. All tasks use rolling windows with stride 1. Normalization statistics are fitted on the training split and reused for validation and testing; MSE and MAE are computed in normalized space. General forecasting models use batch size 32, at most 10 epochs, and early-stopping patience 3, with model-specific hyperparameters tuned within predefined ranges. Agent methods retain the training hyperparameters of their final selected programs and are reevaluated under the same data splits, windows, and metric protocol.

All agent workflows share a formal-improvement cap of at most 20 rounds, with at most three repair attempts per formal round. EvoCast establishes and diagnoses a task-specific baseline before formal rounds, AIDE runs a 10-round draft stage before formal improvement, and R\&D-Agent completes workflow initialization before formal research. Resource totals in Table~\ref{tab:rq2_resource_ledger_main} accumulate from workflow start through termination, including these pre-formal stages, planning, coding, execution, repair, evaluation, and state update. A valid round is a formal round that produces measurable MSE or MAE.

\begin{table*}[t]
\centering
\small
\caption{RQ2 overall forecasting results on three real tasks. Best values are \textbf{bolded} and second-best values are \underline{underlined}. Numbers in parentheses for EvoCast denote relative error reduction against the corresponding selected baseline.}
\label{tab:rq2_overall_results_main}
\begin{tabular}{p{0.24\textwidth}C{0.09\textwidth}C{0.09\textwidth}C{0.09\textwidth}C{0.09\textwidth}C{0.09\textwidth}C{0.09\textwidth}}
\toprule
Method & \multicolumn{2}{c}{Air Quality (MM)} & \multicolumn{2}{c}{Melbourne Pedestrian T33 (SS)} & \multicolumn{2}{c}{Steel Industry (MS)} \\
\cmidrule(lr){2-3}\cmidrule(lr){4-5}\cmidrule(lr){6-7}
 & MSE & MAE & MSE & MAE & MSE & MAE \\
\midrule
Sundial (2025) & 0.5481 & 0.5194 & 0.1814 & 0.2253 & 0.4413 & 0.3997 \\
Toto-2.0-2.5B (2026) & 0.4968 & 0.5061 & 0.2583 & 0.2643 & 0.4880 & 0.3702 \\
Time-VLM (2025) & 0.4737 & \underline{0.4988} & 0.2570 & 0.3067 & 0.3565 & 0.4057 \\
TimeCMA (2025) & 0.5134 & 0.5240 & 1.2620 & 0.9147 & \underline{0.3037} & \underline{0.3658} \\
AIDE (2025) & 0.5309 & 0.5327 & \underline{0.1728} & \textbf{0.2070} & 0.4112 & 0.4780 \\
R\&D-Agent (2025) & 0.5897 & 0.5893 & 0.1833 & 0.2289 & 0.7244 & 0.5420 \\
Selected baseline & \underline{0.4665} & 0.5004 & 0.1852 & 0.2410 & 0.3057 & 0.3733 \\
\raisebox{0.9ex}[0pt][0pt]{EvoCast} & \shortstack{\textbf{0.4570}\\(-2.04\%)} & \shortstack{\textbf{0.4909}\\(-1.90\%)} & \shortstack{\textbf{0.1628}\\(-12.10\%)} & \shortstack{\underline{0.2242}\\(-6.97\%)} & \shortstack{\textbf{0.2792}\\(-8.67\%)} & \shortstack{\textbf{0.3388}\\(-9.24\%)} \\
\bottomrule
\end{tabular}
\end{table*}

\begin{table*}[t]
\centering
\small
\caption{RQ2 valid formal rounds and end-to-end resource ledger. Valid rounds show the ratio and count over 20. Active time is cumulative workflow wall-clock time; Calls and Tokens count all LLM calls and input/output tokens; Cost is total API expenditure. Bold indicates the highest valid-round ratio for each task.}
\label{tab:rq2_resource_ledger_main}
\begin{tabular}{p{0.19\textwidth}lccrrr}
\toprule
Task & Method & Valid rounds $\uparrow$ & Active time (s) $\downarrow$ & Calls $\downarrow$ & Tokens $\downarrow$ & Cost (USD) $\downarrow$ \\
\midrule
Air Quality & AIDE & 60\% (12/20) & 3,436 & 60 & 872,050 & 0.4915 \\
 & R\&D-Agent & 65\% (13/20) & 29,292 & 256 & 7,301,193 & 3.7618 \\
 & EvoCast & \textbf{100\% (20/20)} & 7,989 & 96 & 1,210,702 & 0.6089 \\
\midrule
Melbourne T33 & AIDE & 65\% (13/20) & 11,221 & 60 & 1,127,775 & 1.0902 \\
 & R\&D-Agent & 75\% (15/20) & 14,425 & 148 & 3,445,604 & 2.4359 \\
 & EvoCast & \textbf{90\% (18/20)} & 16,625 & 96 & 1,485,592 & 0.7798 \\
\midrule
Steel Industry & AIDE & 80\% (16/20) & 11,705 & 61 & 843,405 & 0.6973 \\
 & R\&D-Agent & 30\% (6/20) & 15,594 & 192 & 3,067,189 & 1.8685 \\
 & EvoCast & \textbf{90\% (18/20)} & 11,018 & 128 & 1,752,020 & 0.8073 \\
\bottomrule
\end{tabular}
\end{table*}

\begin{figure*}[t]
\centering
\includegraphics[width=\textwidth]{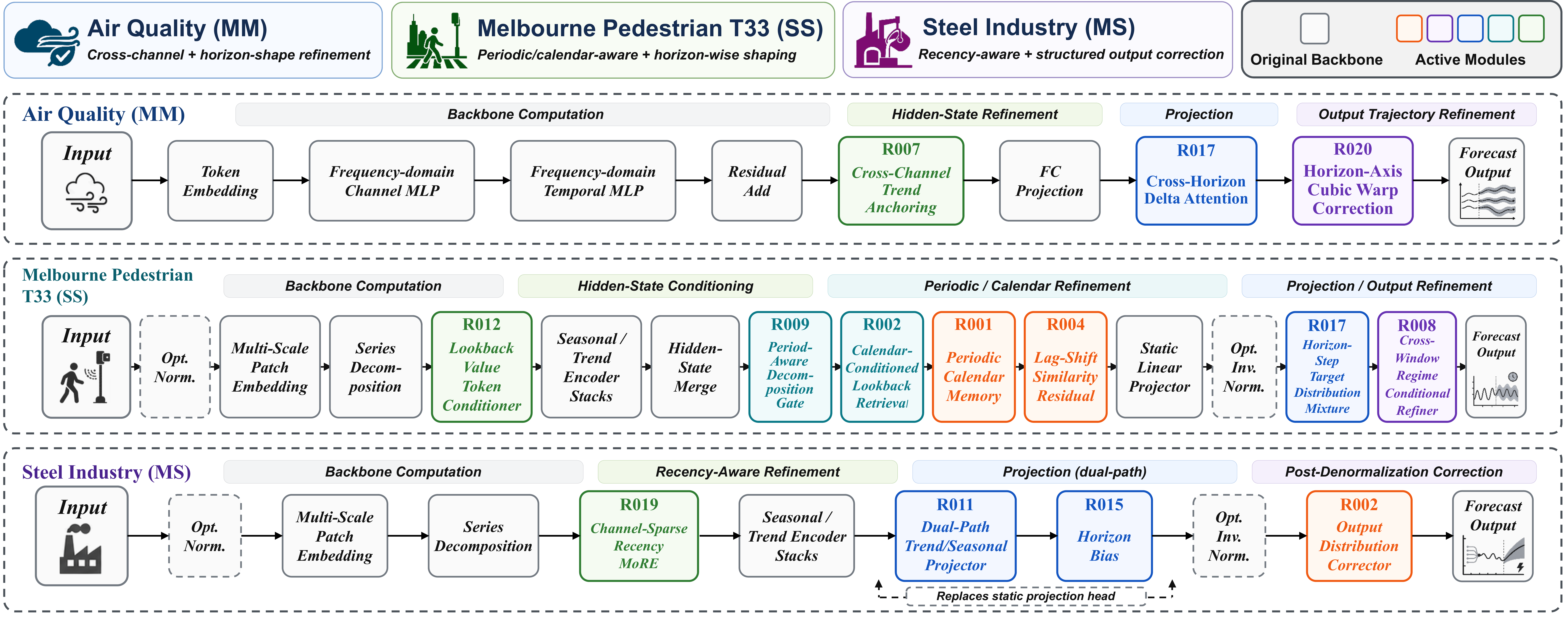}
\caption{Task-adaptive final live-path architectures discovered by EvoCast on the three RQ2 real forecasting tasks. Gray modules are original backbone components, and colored modules are active modules accumulated during EvoCast research rounds.}
\Description{Three model diagrams for Air Quality, Melbourne Pedestrian T33, and Steel Industry. Gray blocks represent inherited backbone components, while colored blocks represent task-specific modules retained after accepted EvoCast research rounds.}
\label{fig:rq2_task_adaptive_architectures}
\end{figure*}

\begin{table*}[t]
\centering
\small
\caption{Overall RQ3 system-ablation results. Final MSE is the three-seed mean over seeds 2026, 2027, and 2028. Lower is better for MSE, time, tokens, and cost; higher is better for valid rounds.}
\label{tab:rq3_ablation_results_main}
\begin{tabular}{p{0.06\textwidth}p{0.27\textwidth}c c c c c}
\toprule
Variant & Removed component & Final MSE & Valid rounds & Time (h) & Tokens (M) & Cost (USD) \\
\midrule
Full & None & \textbf{0.162579} & \underline{18 / 20} & \underline{4.62} & \underline{1.49} & \underline{0.7798} \\
A1 & Dataset characterization & 0.173547 & 17 / 20 & 7.01 & \textbf{1.42} & \textbf{0.6879} \\
A2 & Baseline diagnosis & 0.173473 & 15 / 20 & 6.80 & 2.79 & 1.4694 \\
A3 & Evidence-grounded planning & \underline{0.173250} & \textbf{20 / 20} & 5.07 & 1.88 & 0.9793 \\
A4 & Authority-controlled execution & 0.185001 & 14 / 20 & 10.80 & 25.32 & 7.8792 \\
A5 & Iterative evidence update & 0.178489 & 15 / 20 & \textbf{2.23} & 1.81 & 0.8145 \\
\bottomrule
\end{tabular}
\end{table*}

Table~\ref{tab:rq2_overall_results_main} reports the final formal forecasting results on the three real cases. EvoCast reduces the selected-baseline error on all three tasks. It achieves the lowest MSE on all three tasks and the lowest MAE on two of them. These results show that the fixed EvoCast loop can develop stronger task-specific architectures under one unchanged evaluator, task protocol, and final-artifact evaluation rule.

Table~\ref{tab:rq2_resource_ledger_main} shows that EvoCast achieves the highest valid-round ratio on all three tasks, with lower total tokens and API cost than R\&D-Agent. Compared with AIDE, EvoCast uses more tokens and API cost on Air Quality and Steel Industry, and more tokens but lower API cost on Melbourne T33; their relative wall-clock time varies by task. Together with Table~\ref{tab:rq2_overall_results_main}, these results show that EvoCast obtains lower final MSE and more valid rounds at the reported resource expenditure.

Figure~\ref{fig:rq2_task_adaptive_architectures} visualizes the final live-path architectures retained by EvoCast on the three real forecasting tasks. The gray modules correspond to original backbone components, and the colored modules correspond to task-specific modules that were accumulated during accepted research rounds and retained on the active forecasting path. The three cases lead to different structural landing points, showing that the fixed EvoCast workflow forms different module insertions and path refinements under different task demands.

The final structures reflect task-specific statistical demands: Air Quality develops cross-variable trend anchoring and horizon-shape refinement; Melbourne T33 strengthens periodic memory, calendar-conditioned retrieval, and horizon-wise output shaping; Steel Industry adds recency-aware refinement, dual-path projection, and output-side correction. These cases show how the same workflow retains different effective mechanism combinations from task evidence.

In the Air Quality case, EvoCast keeps and evolves the live path of the selected forecasting backbone, while AIDE converges to a linear Ridge feature-engineering solution and R\&D-Agent to a GRU/LSTM ensemble. This comparison shows the different final artifacts produced by these workflows under the same formal-round cap. Appendix~\ref{app:rq2} provides additional forecasting-model comparisons, trajectories, and architecture details.

\subsection{RQ3: Orchestration and Authority Design}

RQ3 evaluates how EvoCast's system design supports the final outcome. We perform controlled system ablations over the fixed EvoCast workflow, and each variant removes or weakens one key mechanism. The comparison tracks both final forecasting quality and agent-side cost, so this section measures the role of the full research loop.

A1 removes dataset characterization, A2 removes baseline diagnosis, A3 removes evidence-grounded planning, A4 removes authority-controlled execution, and A5 removes iterative evidence update. These variants isolate how task evidence, mechanism diagnosis, planning quality, execution control, and cross-round memory affect the stability of the autonomous research loop.

All variants use Melbourne Pedestrian T33 with PatchMLP as the starting backbone, lookback 168, horizon 24, and MSE as the optimization metric. The formal research cap is 20 rounds, with at most three repair attempts per round. Each variant removes its designated mechanism while preserving the remaining workflow.

Table~\ref{tab:rq3_ablation_results_main} shows that the full loop achieves the lowest final MSE while maintaining controlled resource use. A1 and A2 weaken direction quality; A3 maintains continuous execution but selects weaker directions; A4 simultaneously yields the weakest final quality and the highest resource cost; and A5 loses later-round accumulation without iterative evidence write-back. Each ablation therefore degrades forecasting quality, research stability, or budget efficiency.

\begin{figure}[t]
\centering
\includegraphics[width=\columnwidth]{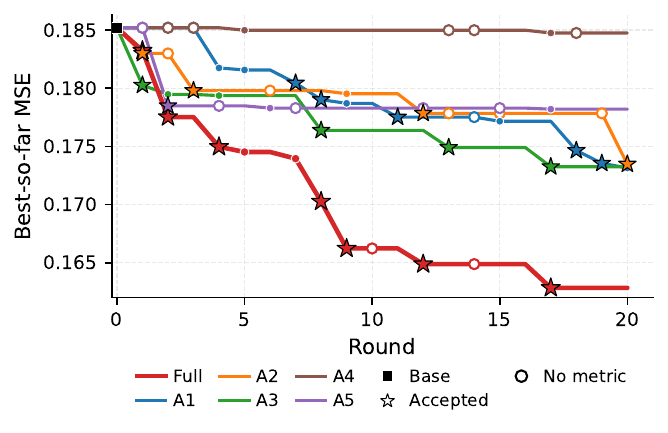}
\caption{RQ3 cumulative best trajectories (event-only). The plot records only events that refresh the best-so-far result, showing when each ablation chain finds a new best candidate and whether later rounds continue to improve.}
\Description{Line chart comparing event-only cumulative-best trajectories for the full EvoCast system and five ablations. The full system reaches a stronger error range earlier and records later best-result updates, while several ablations plateau sooner.}
\label{fig:rq3_event_only}
\end{figure}

Figure~\ref{fig:rq3_event_only} further shows the cumulative best trajectories of the ablation chains. Because the plot records only best-refreshing events, it directly reveals when each chain reaches a new best candidate and whether later rounds continue to make progress. The full EvoCast chain enters a stronger error region earlier and continues to accumulate later improvements. When key system mechanisms are weakened, the trajectories stagnate earlier or yield only limited local gains. This evidence shows that the RQ3 differences appear in both the final score and the system's ability to sustain effective research progress across rounds. Appendix~\ref{app:rq3} details the component ablations.

\section{Conclusion}
\label{sec:conclusion}

EvoCast turns forecasting architecture development into an autonomous process that can establish a task-specific starting point, accumulate executable research evidence across rounds, and deliver a final forecasting architecture tailored to the target scenario. Along this workflow, the system establishes a task-specific baseline, constructs mechanism evidence, advances live-path architecture evolution, and ultimately develops a forecasting model suited to the current scenario. Each accepted architecture change is tied to executable source code, canonical evaluation, and a traceable state update, so the final result is both reusable and practically deployable.

The main value of this work is that it advances autonomous time-series modeling toward task-specific model development. For forecasting research, this means that one fixed workflow can develop suitable time-series models for different real-world scenarios. More broadly, it shows that the central value of autonomous scientific coding systems lies in producing executable, auditable, and scenario-adaptive research artifacts around concrete tasks.

\bibliographystyle{ACM-Reference-Format}
\bibliography{references}

\clearpage
\input{appendix}

\end{document}

%% file: appendix.tex
\appendix
\setcounter{secnumdepth}{3}
\counterwithin{figure}{section}
\counterwithin{table}{section}
\counterwithin{algorithm}{section}

\section{Extended Related Work}
\label{app:related-work}

\subsection{Deep Time-Series Forecasting Architectures}

Deep time-series forecasting has developed into a heterogeneous architecture ecosystem. Representative families cover decomposition and frequency modeling, patch-level temporal representation, cross-variate dependency modeling, linear modeling, dense encoding, and time-frequency hybrid modeling \citep{wu2021autoformer,zhou2022fedformer,nie2023patchtst,zhang2023crossformer,liu2024itransformer,zeng2023are,das2023tide,wu2023timesnet}. These architecture families encode different assumptions about temporal structure, variable interaction, and information flow. TSLib systematizes this ecosystem as a reusable benchmark library for deep time-series models across forecasting, imputation, anomaly detection, and classification \citep{wang2024deep}.

This architectural diversity creates a direct need for task-specific development. TFB further provides a controlled and fair evaluation protocol for forecasting and shows that forecasting architectures exhibit strongly dataset-dependent behavior \citep{qiu2024tfb}. These observations make baseline selection, controlled comparison, and architecture modification central parts of forecasting research. EvoCast builds on this setting and organizes baseline establishment, baseline diagnosis, and architecture evolution within one fixed-protocol research loop.

\subsection{Automated Forecasting Model and Architecture Development}

Automated model development has been studied extensively through AutoML, hyperparameter optimization, and neural architecture search. AutoML and hyperparameter optimization automate model selection and configuration, while NAS explores candidate structures inside predefined operator spaces or architecture templates \citep{he2021automl,elsken2019nas}. These methods provide strong automation within explicit search spaces.

Recent LLM-assisted architecture design extends automation from fixed search spaces to dynamic structure construction. NADER organizes architecture design as a multi-agent process, AgentNAS generates a seed architecture and then constructs a bounded task-specific search space around it, and GenAutoML generates task-specific forecasting architectures together with runtime checks and self-repair \citep{yang2024nader,jeong2026agentnas,poikarayil2026genautoml}. These methods expand automated architecture development from static search to open-ended structure generation.

EvoCast focuses on repository-bound forecasting architecture development. The development target is executable source modules inside a fixed forecasting repository, and every modification is constrained by the protocol, source boundary, and canonical evaluator. EvoCast starts from an established forecasting repository baseline that is already bound to a fixed task protocol and evaluator, forms evidence-grounded module-level research directions on real source code, and advances architecture evolution through bounded repository-level modification. This setting requires the system to understand active mechanisms, execution paths, and protocol constraints before proposing new architectural changes and to write each round outcome into a reusable evidence state.

\begin{table*}[t]
\centering
\small
\caption{Comparison of representative time-series agent systems.}
\label{tab:rw_timeseries_agents}
\begin{tabular}{@{}L{0.16\textwidth}L{0.17\textwidth}L{0.17\textwidth}L{0.23\textwidth}L{0.17\textwidth}@{}}
\toprule
System & Starting point & Modification target & Evidence for next step & Promotion authority \\
\midrule
TimeSeriesScientist \citep{zhao2025timeseriesscientist} & forecasting workflow and model pool & workflow and model configuration & workflow observations and validation results & workflow-level selection \\
SEA-TS \citep{xu2026seats} & searched forecasting code state & generated forecasting code & search feedback, review results, and execution outcomes & execution-result-based selection \\
GenAutoML \citep{poikarayil2026genautoml} & generated architecture scaffold & task-specific model architecture & runtime checks and repair feedback & runtime-validation-based selection \\
EvoCast & task-specific repository baseline & repository source modules & executed baseline ablations, prior-round evidence, and failure memory & deterministic promotion gate outside the LLM \\
\bottomrule
\end{tabular}
\end{table*}

\subsection{LLMs and Agentic Systems for Time Series}

LLM and agentic methods for time series first emerged in inference-time forecasting, representation learning, and reasoning. Time-LLM repurposes large language models as forecasting backbones, TimeCMA improves multivariate forecasting through cross-modality alignment between time-series and prompt embeddings, Time-VLM integrates temporal, visual, and textual representations in a multimodal forecasting framework, TS-Agent iteratively collects statistical evidence for time-series reasoning, and Nexus coordinates specialized agents over temporal patterns and contextual information \citep{jin2024timellm,liu2025timecma,zhong2025timevlm,liu2025tsagent,das2026nexus}. These systems place foundation models inside the forecasting or reasoning workflow.

Another line of work automates forecasting workflows. TimeSeriesScientist automates data understanding, planning, validation, and reporting for time-series analysis \citep{zhao2025timeseriesscientist}. This line demonstrates that forecasting workflows can be organized by agent systems at the task level.

The closest systems to EvoCast operate at the level of forecasting algorithm or forecasting architecture generation. SEA-TS formulates forecasting code generation as a self-evolving process with search, review, and historical feedback, and GenAutoML focuses on task-specific architecture generation and repair \citep{xu2026seats,poikarayil2026genautoml}. EvoCast combines executable task-specific baseline establishment, executed baseline diagnosis, persistent evidence state, and deterministic promotion authority outside the LLM. This combination supports fixed-protocol repository-level architecture evolution across rounds.

Table~\ref{tab:rw_timeseries_agents} highlights the combination that defines EvoCast: a fixed repository baseline, executed diagnostic evidence, persistent evidence state, repository-level modification, and deterministic promotion authority.

\subsection{Autonomous Research Agents and Authority-Controlled Experimentation}

Autonomous research agents already cover hypothesis generation, code editing, experiment execution, and long-horizon iteration. MLAgentBench provides a standardized environment for studying agent performance on machine learning experimentation, The AI Scientist and The AI Scientist-v2 extend automation to idea generation, experimentation, and paper writing, InternAgent builds a closed loop from hypothesis to verification, and R\&D-Agent separates research and development roles \citep{huang2024mlagentbench,lu2024aiscientist,yamada2025aiscientistv2,zhang2025internagent,yang2025rdagent}. These systems establish the broader context for autonomous ML research.

MLE-STAR is especially relevant because it uses executed ablations to identify high-impact code regions and then applies targeted refinement \citep{nam2025mlestar}. This design shows that execution-grounded ablation can directly guide later research decisions. EvoCast carries this principle into forecasting architecture research and focuses the ablation target on active baseline mechanisms, so diagnostic outcomes become task-specific evidence for later planning.

Reliable experimentation infrastructure further determines whether autonomous research conclusions are auditable. ResearchLoop organizes task contracts, evidence objects, and persistent project state in one control plane, AIRA\textsuperscript{2} emphasizes hidden consistent evaluation in long-horizon research, FML-bench and ResearchGym separate agent strategy from execution infrastructure, and Arbor maintains cross-round hypotheses, artifacts, and evidence through hypothesis-tree refinement \citep{xia2026researchloop,hambardzumyan2026aira2,zou2026fmlbench,garikaparthi2026researchgym,jin2026arbor}. EvoCast instantiates these principles in forecasting architecture development by assigning program authorities to source boundaries, canonical evaluation, promotion, result semantics, and persistent state updates.

\section{System and Implementation Details}
\label{app:system}

This section expands the fixed-protocol autonomous architecture development loop described in Section~\ref{sec:method}. Given a forecasting task, a model repository, and a bounded research budget, EvoCast fixes the task protocol, establishes a comparable baseline, converts the baseline into executed mechanism evidence, runs bounded architecture research rounds, updates the research state through program authorities, and generates a final HTML report from frozen facts.

The goal of this section is to explain the execution semantics behind Section~\ref{sec:method}. Cognitive agents generate ablation targets, research directions, candidate source states, and final narrative. Program authorities control source boundaries, canonical evaluation, promotion decisions, result semantics, and state write-back. This division keeps open-ended architecture exploration inside a fixed comparison surface.

\subsection{Detailed Workflow}

\paragraph{End-to-End Execution Path.}

EvoCast organizes one autonomous research task as a fixed execution path. The system first reads the task card and fixes the dataset, forecasting mode, targets, lookback, horizon, metric, training budget, research-round budget, and repair budget. It then establishes the active baseline under the fixed protocol and performs mechanism-level diagnosis on that baseline. During formal research, the Research Planner reads task evidence, mechanism evidence, prior-round evidence, failure memory, and the current best model to select one module-level direction. The Coding / Builder Agent implements that direction within a bounded source scope, and the Program Authority Chain performs source validation, canonical evaluation, promotion judgment, and state update. After termination, the Report Narrative Agent reads frozen facts and generates the final HTML report.

The execution path is grounded in a fixed comparison surface. All baselines, ablation candidates, research candidates, and seed evaluations share the same task protocol. Each round therefore has a unified interpretation: accepted candidates update the model state, valid rejections update scientific evidence, unstable improvements become weak positive evidence, and engineering failures update failure memory. The next planner invocation reads the updated state, which turns the loop into evidence-driven iteration.

Figure~\ref{fig:method_workflow} shows the complete runtime path of EvoCast. The upstream stages fix the task protocol and establish the baseline, the middle stages perform baseline diagnosis and formal research rounds, and the downstream stages handle authoritative adjudication, evidence-carrying state update, and final reporting. The central relation is the closed-loop state flow: accepted results, negative results, repair records, and failure records are written into the research state and become inputs to later Research Planner invocations.

\paragraph{Cognitive Roles and Program Authorities.}

EvoCast separates cognitive roles from program authorities. The cognitive roles include the Baseline-Diagnosis Planner, the Research Planner, the Coding / Builder Agent, and the Report Narrative Agent. They generate ablation targets, research directions, candidate source states, and final report narrative. Program authorities control source boundary checking, candidate materialization, canonical evaluation, promotion gate decisions, result semantics, and persistent state update.

This separation defines three permissions. LLM agents can generate research intent and code changes. Builder edits must remain inside the fixed editable boundary. Experimental facts, metrics, promotion decisions, and model-state updates are confirmed only by program authorities. Agent self-report does not directly become an experimental conclusion.

\paragraph{Boundary Enforcement.}

Boundary enforcement separates model-side editing from evaluation-side fact production. The editable region covers only the model-side source files allowed in the current round. The dataset loader, split definition, metric computation, training schedule, evaluation runner, and promotion logic remain frozen outside the Builder's editable scope. Each candidate first passes source-boundary validation before canonical evaluation is launched. Candidates that violate the frozen protocol are marked as protocol-invalid results and do not enter formal comparison.

Figure~\ref{fig:method_role_authority} expands the cognition-authority boundary inside one controlled research round. The cognitive layer proposes directions, implements candidates, and generates narrative. The authority layer checks candidate scope, runs the canonical evaluator, applies the promotion rule, and writes model and evidence states. This boundary explains how EvoCast keeps open-ended architecture search under deterministic execution control.

\subsection{Fixed Task Protocol and Task Grounding}

\paragraph{Task and Protocol Initialization.}

Each autonomous run starts from a fixed task protocol. Before research begins, the system fixes the dataset, target, lookback, horizon, forecasting mode, metric, training budget, research-round budget, and repair budget. The protocol is written into the grounded task package and is reused by baseline evaluation, ablation execution, candidate evaluation, and seed-stability checks.

The fixed protocol defines the comparison surface for the entire autonomous run. The Research Planner preserves the task definition. The Coding / Builder Agent preserves the training strategy, input length, prediction horizon, and evaluation rule. The Program Authority Chain generates formal results under the same metric and evaluator. Improvements, rejections, and failures are therefore interpreted on one shared semantic surface.

\paragraph{Dataset Characterization as Task Evidence.}

Task grounding also converts the raw dataset into planner-readable evidence. EvoCast follows the feature-computation protocol of the TFB benchmark \citep{qiu2024tfb} and extracts correlation, transition behavior, distribution shift, seasonality, trend, stationarity, and short- and long-term distribution differences from valid variables.

For a univariate dataset, the system keeps the feature vector of the target variable. For a multivariate dataset, it aggregates valid variables to form a dataset-level profile:
\[
\begin{aligned}
\mathbf z_D =
\bigl[
&z_{\mathrm{corr}},
\bar z_{\mathrm{trans}},
\bar z_{\mathrm{shift}},
\bar z_{\mathrm{season}},
\bar z_{\mathrm{trend}},\\
&\bar z_{\mathrm{station}},
\bar z_{\mathrm{short}},
\bar z_{\mathrm{long}}
\bigr].
\end{aligned}
\]

These features serve only as planning evidence. They describe whether the task exhibits cross-variable dependency, distribution drift, periodic structure, trend, or nonstationarity, and they are read together with baseline diagnosis, prior rounds, and failure memory. They do not directly choose a specific architecture modification.

\paragraph{Baseline Establishment.}

Baseline establishment converts the fixed task into a comparable architecture starting point. Under the fixed protocol, EvoCast runs baseline candidates and selects the active baseline through canonical evaluation. This baseline serves as both the comparison anchor for later research candidates and the executed starting point for baseline diagnosis.

This stage fixes the source base for architecture evolution. Ablation targets, research directions, and candidate comparisons are all defined around the active baseline or the promoted current best model.

\subsection{Baseline Establishment and Diagnosis}

\paragraph{Executed Mechanism Ablation.}

Baseline diagnosis converts the active baseline from a comparison anchor into mechanism evidence. EvoCast first analyzes the active prediction path and then uses the Baseline-Diagnosis Planner to generate runnable mechanism-ablation targets. Each ablation target is tied to a concrete source region, an ablation action, and an input-output preservation contract.

Each ablation target goes through the same execution chain used in a formal research round. The Coding / Builder Agent implements the ablation candidate, and the Program Authority Chain performs validation, canonical evaluation, and adjudication. Only ablations that produce valid evaluation results enter later research evidence.

This execution policy gives baseline diagnosis formal experimental status. Mechanism importance is determined by real execution under the shared protocol.

\paragraph{Outcome Semantics of Ablation.}

Ablation outcomes are converted into planner-readable diagnostic semantics. Clear degradation marks the removed mechanism as essential for the current task. Clear improvement marks the mechanism as harmful or redundant and makes it a replacement or restructuring target. Small change marks the mechanism as weak or uncertain. Build/runtime failures enter engineering failure memory, and protocol violations are excluded from formal research evidence.

\begin{table*}[t]
\centering
\small
\caption{Diagnostic semantics of ablation outcomes.}
\label{tab:s1_ablation}
\begin{tabular}{lll}
\toprule
Ablation result & Diagnostic label & Planning consequence \\
\midrule
Clear degradation & essential & Preserve the core function and modify cautiously \\
Clear improvement & harmful or redundant & Prioritize replacement or restructuring \\
Small change & weak or uncertain & Keep it as an exploration target \\
Build/runtime failure & engineering failure & Do not draw a scientific conclusion \\
Protocol violation & invalid & Discard or rerun \\
\bottomrule
\end{tabular}
\end{table*}

Table~\ref{tab:s1_ablation} shows how executed ablation results enter the Research Planner. The table converts ablation outcomes into preserved mechanisms, replacement targets, uncertain targets, and failure memory, so later direction selection is grounded in executed evidence.

\subsection{Evidence-Grounded Research Direction Selection}

\paragraph{Role of the Research Planner.}

The Research Planner runs after baseline diagnosis. It reads the grounded task package, active baseline or current best source structure, ablation evidence, positive and negative prior rounds, repair history, and failure memory, and then selects one module-level research direction. Each round advances exactly one direction so that later results, rejections, and failures can be attributed clearly.

To stay consistent with the notation in Section~\ref{sec:preliminaries}, the round-level research state read by the planner is denoted by
\[
\mathcal{S}_r = (M_r, E_r, \mathcal{H}_r),
\]
where $M_r$ is the current-best model state at the start of round $r$, $E_r$ is the accumulated scientific and engineering evidence state, and $\mathcal{H}_r$ is the structured history that stores prior rounds, repair traces, failure records, and gate decisions. This definition makes the shared research state explicit before later planner, authority, and state-update steps reuse it.

\begin{figure*}[t]
\centering
\fbox{%
\parbox{0.96\textwidth}{%
\small
\textbf{Planner Prompt Summary.}\\[2pt]
\textbf{SYSTEM}\\
\ttfamily You are an autonomous time-series model research planner. Your task is to choose the next research idea from dataset analysis, ablation evidence, prior rounds, failure memory, model structure, and task constraints. Select exactly one idea and return it as structured JSON.\\[4pt]
\textbf{USER}\\
\ttfamily planner\_architecture: research\_program\_hypothesis\_competition\\
\ttfamily architecture\_instruction: infer the unresolved architectural need from dataset facts, ablation evidence, prior rounds, failure memory, and model structure; then generate multiple standalone module-level ideas and select one direction that changes information flow while preserving the fixed forecasting protocol.\\[2pt]
\ttfamily hard\_requirements:\\
\ttfamily (1) Use dataset, ablation, and prior-round evidence together.\\
\ttfamily (2) Treat prior rounds as evidence, not as a mechanism library for local continuation.\\
\ttfamily (3) Select a standalone module-level experiment, not the nearest local variant.\\
\ttfamily (4) Build from the active baseline source, not from prior unpromoted variants.\\
\ttfamily (5) Do not treat build, import, shape, evaluator, provenance, or runtime failures as scientific evidence.\\
\ttfamily (6) Do not judge idea quality by expected metric improvement alone.\\
\ttfamily (7) Reject bias-only, gate-only, dropout-only, norm-only, activation-only, or parameter-only tweaks.\\
\ttfamily (8) Reject wrappers that do not create a new information path.\\
\ttfamily (9) Do not propose dataset, metric, optimizer, scheduler, epoch, batch size, or learning-rate changes.\\
\ttfamily (10) Return one executable direction with a target mechanism, source region, expected behavior change, failure to avoid, and expected knowledge gain.\\[2pt]
\ttfamily required\_json\_shape: idea\_title, hypothesis, target\_mechanism, target\_code\_region, expected\_behavior\_delta, failure\_to\_avoid, what\_this\_round\_will\_teach.\\
\ttfamily evidence\_bundle\_text: <runtime task evidence>.%
}}
\caption{Planner prompt summary used for research-direction generation. Low-value metadata and repeated requirements are omitted, while the two-stage planning logic, the evidence semantics, the anti-local-continuation constraints, and the structured output fields are preserved.}
\Description{Structured prompt summary for the research planner. It contains task and baseline evidence, history, failure memory, a two-stage planning procedure, constraints, and required output fields for a research direction.}
\label{fig:s4_planner_prompt}
\end{figure*}

The output is a structured research direction containing the target mechanism, target source region, testable hypothesis, expected information-flow change, implementation risk, and expected knowledge gain. The planner decides what mechanism to study, why the current evidence supports that study, and what the round should teach. It does not implement code, run experiments, or update the current best model.

Formally, the planner output remains
\[
d_r = \Pi(\mathcal{P}, \mathcal{S}_r),
\]
where $\Pi(\cdot)$ denotes the direction-generation policy. For later execution, the direction can be written as
\[
d_r = (m_r, \rho_r, h_r, \kappa_r),
\]
where $m_r$ denotes the target mechanism, $\rho_r$ denotes the target code region, $h_r$ denotes the testable hypothesis, and $\kappa_r$ denotes the structured summary of implementation risk and expected knowledge gain. This decomposition clarifies that $d_r$ is a structured research intent that can be consumed by the Builder and the authority-controlled execution path.

Figure~\ref{fig:s4_planner_prompt} gives the planner prompt as an audit artifact for research-direction generation.

\paragraph{Evidence Bundle for Planning.}

The Research Planner rereads a unified evidence bundle at every round. Its input can be compressed into four evidence groups: task evidence, model evidence, scientific evidence, and engineering evidence. Task evidence describes data structure, horizon, variable scale, and protocol boundaries. Model evidence describes the baseline, current best, and source structure. Scientific evidence records ablations, accepted candidates, valid rejections, and unstable improvements. Engineering evidence records import, shape, training, repair, and source-boundary failures.

\begin{table*}[t]
\centering
\small
\caption{Planner evidence bundle.}
\label{tab:s2_planner_evidence}
\begin{tabular}{lll}
\toprule
Evidence group & Examples & Planning use \\
\midrule
Task evidence & data features, horizon, variable count & identify task needs \\
Model evidence & baseline, current best, source structure & determine the starting point \\
Scientific evidence & ablations, positive results, negative results & update mechanism judgment \\
Engineering evidence & import, shape, training, repair history & identify implementation risks \\
\bottomrule
\end{tabular}
\end{table*}

Table~\ref{tab:s2_planner_evidence} gives the minimal planner input set. Engineering evidence is used to identify implementation risks. Scientific evidence updates mechanism judgment. This distinction prevents build, shape, provenance, and runtime failures from being interpreted as mechanism-level conclusions.

\paragraph{Planning Principle.}

The core job of the Research Planner is to convert accumulated evidence into the next testable hypothesis. It first identifies which architecture capability is still missing for the current task, then compares multiple mechanism-level candidates, and finally keeps one direction for the round. A valid direction must be grounded in the current evidence, map to a real source region of the active architecture, and increase later research knowledge whether the outcome is positive, negative, unstable, or failed.

The resulting research direction contains four categories of content. The research objective states what the round studies and why now is the right time to study it. The testable hypothesis defines the expected information-flow change and the empirical outcome that would support or weaken it. The implementation guidance points to the target source region and implementation risks. The expected knowledge gain explains how the round will affect later planning. These four parts define the research intent but do not possess execution authority.

\paragraph{Authority Boundary.}

The Research Planner has direction-generation authority. It can interpret evidence, compare hypotheses, select a source-grounded direction, and define expected knowledge gain. It cannot modify source code, change the task protocol, expand the editable boundary, run evaluation, or update the current best model. The next stages handle implementation and experimental adjudication.

\subsection{Candidate Construction and Program Authorities}
\label{sec:system-overview-execution}

\paragraph{Role of the Coding / Builder Agent.}

\begin{figure*}[t]
\centering
\fbox{%
\parbox{0.96\textwidth}{%
\small
\textbf{Builder Prompt Summary.}\\[2pt]
\textbf{SYSTEM}\\
\ttfamily You are implementing one research model source edit using a complete-file workflow. Return exactly one JSON object. Produce complete replacement content for the edited allowed file. Preserve the public model interface, tensor-shape contract, task semantics, input length, prediction horizon, target selection, and training hyperparameters. Activate the proposed mechanism in the effective forecast path. Do not edit data, metrics, runner, config, or evaluation code.\\[4pt]
\textbf{USER}\\
\ttfamily task: generate a complete-file implementation for this bounded execution contract.\\[2pt]
\ttfamily required\_output\_schema:\\
\ttfamily files: path/to/allowed\_file.py $\rightarrow$ complete replacement UTF-8 source code.\\
\ttfamily metadata: display\_idea, idea\_summary, mechanism\_name, changed\_mechanism, expected\_effect.\\
\ttfamily activation\_evidence: short evidence that the new mechanism is instantiated and called in the live forecast path.\\[2pt]
\ttfamily path\_policy: existing files may modify only allowed edit files; new files may be created only under allowed new-file roots.\\[2pt]
\ttfamily execution\_contract:\\
\ttfamily implementation\_goal: mechanism\_name, target\_code\_region, required\_change, hypothesis, expected\_effect.\\
\ttfamily editable scope: primary\_edit\_files, allowed\_edit\_files, allowed\_new\_file\_roots.\\
\ttfamily task semantics: dataset, task\_mode, objective\_metric, runtime\_hyper\_params.\\
\ttfamily shape\_contract: preserve runtime task semantics, channel dimensions, runtime output shape, and callable class interfaces.\\
\ttfamily avoid: do not change dataset, metric, evaluation semantics, horizon, input length, batch size, epoch count, patience, or optimizer schedule; do not write outside allowed edit files or allowed new-file roots.\\
\ttfamily execution\_workflow: start with primary edit files; after the first successful edit, run internal checks and finish with metadata; if the mechanism is source-infeasible, return precise infeasibility metadata.\\[2pt]
\ttfamily source\_files: <allowed file path> $\rightarrow$ <current full source code>.\\
\ttfamily previous\_failed\_attempts: <recent failure records>.\\
\ttfamily repair\_message\_from\_orchestrator: <repair message>.%
}}
\caption{Builder prompt summary used for bounded candidate implementation. The complete-file workflow, the execution-contract fields, the path and shape constraints, the live-path activation requirement, and the repair inputs are preserved, while repeated low-level fields are omitted.}
\Description{Structured prompt summary for the builder agent. It specifies the BuildContract, allowed source paths, implementation constraints, live-path activation requirement, canonical evaluation requirements, and repair feedback.}
\label{fig:s5_builder_prompt}
\end{figure*}

The Coding / Builder Agent converts the selected research direction into a candidate source state. It reads the research intent, source context, allowed edit files, allowed new-file roots, shape contract, and repair feedback, and returns replacement source content through a complete-file workflow. The Builder must place the proposed mechanism on the effective forecast path while preserving the public model interface, tensor-shape contract, task semantics, input length, prediction horizon, target selection, and training hyperparameters.

Figure~\ref{fig:s5_builder_prompt} gives the builder prompt summary used for bounded candidate implementation.

The Builder is responsible for implementation. It cannot modify the dataset, metric, runner, configuration, training budget, evaluation semantics, or editable boundary, and it cannot authorize a performance conclusion. Performance conclusions are produced only by canonical evaluation and the Promotion Gate.

\paragraph{Program Authority Chain.}

The Program Authority Chain converts the Builder output into a formal experimental conclusion. It contains source boundary authority, candidate materialization authority, canonical evaluation authority, promotion authority, and state-update authority. Candidate source code first passes source compliance and artifact construction, then runs through the fixed evaluator to generate canonical metrics, and finally enters the Promotion Gate.

Consistent with Section~\ref{sec:method}, this stage corresponds to a two-stage authority-controlled execution path:
\[
c_r = \beta(d_r, \mathcal{P}, M_r),
\]
\[
o_r = \Gamma(\mathcal{P}, \mathcal{S}_r, c_r).
\]
Here $c_r$ denotes the canonical candidate that enters formal comparison and $o_r$ denotes the formal round outcome. $\beta(\cdot)$ denotes candidate construction under the fixed protocol and the current research state, while $\Gamma(\cdot)$ denotes the program-authority process that applies boundary checking, canonical evaluation, and promotion adjudication. This split separates research-direction generation, candidate construction, and formal experimental conclusion.

The candidate-execution chain proceeds from the Builder's final workspace to candidate materialization, external verification, canonical evaluation, promotion-gate judgment, and evidence plus model-state update.

Promotion Gate is the deterministic adjudication stage for model-state updates. It handles candidates that have passed source validity and runtime validity, compares them against the current best under the fixed metric and comparison rule, and applies the seed-stability requirement before authorizing promotion.

Single-run improvement is recorded as evidence, while stable improvement is required for model-state promotion. EvoCast places seed evaluation inside the Promotion Gate so that current-best updates depend on reproducible gain. This stability check protects later planning, evidence accumulation, and final reporting from incidental seed effects.

Promotion semantics are therefore split into two levels. The first level is evidence recording: single-run improvements, valid rejections, and unstable gains can all be written back as executed evidence. The second level is incumbent replacement: only candidates that pass source-boundary validation, produce valid canonical metrics, and satisfy the fixed multi-seed comparison rule can update the current best model. This split keeps evidence accumulation and model-state promotion distinct.

This authority chain turns a candidate source state into an experiment with clear research semantics. The research conclusion comes from real execution and fixed rules.

\paragraph{Execution Semantics.}

The Program Authority Chain distinguishes source validity, runtime validity, and scientific admissibility. Source validity checks whether the candidate respects the editable boundary and forms a formal candidate. Runtime validity checks whether the candidate runs under the shared evaluator. Scientific admissibility checks whether the result satisfies the predefined comparison and stability rules. Only candidates that pass all three levels can affect the model state.

Under this framework, candidate execution yields four formal outcome classes: accepted candidate, valid rejection, unstable improvement, and engineering failure. An accepted candidate updates the current best model. A valid rejection produces confirmed negative evidence. An unstable improvement becomes weak positive evidence and does not update the current best. An engineering failure updates failure memory and constrains later implementation paths.

Formally, the round outcome is written as
\[
o_r \in
\{
\mathrm{accepted},
\mathrm{rejected},
\mathrm{unstable},
\mathrm{failed}
\}.
\]
This notation aligns the execution semantics with the gate decision in Algorithm~\ref{alg:s1_round_lifecycle} and with the outcome-specific write-back semantics defined later in Table~\ref{tab:s3_state_updates}.

\subsection{Iterative State Update and Final Reporting}

\paragraph{One-Round Lifecycle.}

One formal research round contains five steps: research-direction selection, candidate implementation, authoritative execution, outcome classification, and state update.

Each round focuses on exactly one research direction and closes with a recordable state update. A round can produce a positive result, negative result, unstable improvement, or engineering failure. All four outcomes carry later planning semantics.

To remain fully aligned with Section~\ref{sec:method}, the algorithm reads the round state through
\[
\mathcal{S}_r = (M_r, E_r, \mathcal{H}_r),
\qquad
d_r = \Pi(\mathcal{P}, \mathcal{S}_r).
\]

\begin{algorithm}[t]
\caption{Lifecycle of one autonomous research round.}
\label{alg:s1_round_lifecycle}
\small
\begin{algorithmic}[1]
\REQUIRE $M_r, E_r, \mathcal{P}, B_{\text{repair}}, d_r$
\ENSURE $M_{r+1}, E_{r+1}, R_r$
\STATE Initialize workspace $W_r$
\STATE Send $d_r$, $M_r$, and protocol constraints $\mathcal{P}$ to the Coding / Builder Agent
\STATE $\mathrm{attempt} \leftarrow 0$
\WHILE{$\mathrm{attempt} \leq B_{\text{repair}}$}
\STATE $S_r \leftarrow$ Builder generates or repairs candidate source code in $W_r$
\STATE $Q_r, F_r, C_r \leftarrow \textsc{ExecuteCandidate}(S_r, \mathcal{P})$
\IF{$Q_r$ is valid}
\STATE \textbf{break}
\ENDIF
\IF{$\mathrm{attempt} = B_{\text{repair}}$}
\STATE $R_r \leftarrow \textsc{EngineeringFailure}(d_r, F_r)$
\STATE \textbf{return} $M_r, \textsc{UpdateState}(E_r, R_r), R_r$
\ENDIF
\STATE Return structured failure feedback $F_r$ to the same Builder session
\STATE $\mathrm{attempt} \leftarrow \mathrm{attempt}+1$
\ENDWHILE
\STATE $G_r \leftarrow \textsc{CompareAndSeedGate}(Q_r, C_r, M_r, \mathcal{P})$
\IF{$G_r = \textsc{accepted}$}
\STATE $R_r \leftarrow \textsc{AcceptedCandidate}(d_r, C_r, Q_r)$
\STATE $M_{r+1} \leftarrow \textsc{Promote}(M_r, C_r, G_r)$
\ELSIF{$G_r = \textsc{rejection}$}
\STATE $R_r \leftarrow \textsc{ValidScientificRejection}(d_r, C_r, Q_r)$
\STATE $M_{r+1} \leftarrow M_r$
\ELSE
\STATE $R_r \leftarrow \textsc{UnstableImprovement}(d_r, C_r, Q_r)$
\STATE $M_{r+1} \leftarrow M_r$
\ENDIF
\STATE $E_{r+1} \leftarrow \textsc{UpdateState}(E_r, R_r)$
\STATE \textbf{return} $M_{r+1}, E_{r+1}, R_r$
\end{algorithmic}
\end{algorithm}

For clarity, Algorithm~\ref{alg:s1_round_lifecycle} uses a compact notation for the round state and execution artifacts. $M_r$ denotes the model state at round $r$, $E_r$ denotes the accumulated evidence state, $\mathcal{P}$ denotes the fixed task protocol, and $d_r$ denotes the research direction selected for the current round. Inside the round, $W_r$ is the round workspace, $S_r$ is the candidate source state produced by the Builder, $Q_r$ is the formal evaluation result, $F_r$ is structured failure feedback, $C_r$ is the canonical candidate artifact, $G_r$ is the gate decision, and $R_r$ is the final round record. These local symbols keep the algorithm compact, and they correspond to the unified research state and authority-controlled execution defined in Section~\ref{sec:method}.

Algorithm~\ref{alg:s1_round_lifecycle} keeps only the main control flow of one research round. The internal details of the authority chain are explained in Section~\ref{sec:system-overview-execution}, repair semantics are explained in Section~\ref{sec:system-overview-repair}, and the state-update semantics of different outcomes are defined later in this section. The \textsc{CompareAndSeedGate} stage includes the seed-stability check used for promotion decisions. The algorithm therefore explains how a candidate enters authoritative execution, how failure is repaired within the same round, and how the round closes into accepted candidate, valid rejection, unstable improvement, or engineering failure.

\paragraph{Repair within the Same Round.}
\label{sec:system-overview-repair}

When a candidate fails during verification or formal evaluation, the system returns structured failure feedback to the same Builder session inside the current round and allows repair within the repair budget. Repair keeps the same research direction, task protocol, editable boundary, and comparison rule.

This design confines engineering repair to candidate executability restoration. The Builder can use runtime feedback to fix import, shape, interface, or live-path activation issues, while the research goal and modification scope remain fixed. This keeps round attribution clear and reduces repeated restarts.

\paragraph{Model State and Evidence State.}

EvoCast maintains two core state types. The model state records the current baseline, current best model, comparison anchor, and promotion history. The evidence state records ablation outcomes, historical directions, valid negative results, unstable improvements, accepted candidates, repair records, and engineering failures. State updates are separated by outcome type: accepted candidates update the model state; valid rejections and unstable improvements update scientific evidence; engineering failures update failure memory; repair history updates implementation-risk evidence.

Consistent with Section~\ref{sec:method}, incumbent replacement is written as
\[
M_{r+1} =
\begin{cases}
c_r, & \text{if } o_r = \mathrm{accepted}, \\
M_r, & \text{otherwise}.
\end{cases}
\]
This rule covers only the current-best model update. Accepted candidates replace the incumbent, while the other outcomes keep the incumbent unchanged.

The evidence update is written as
\[
E_{r+1} = \Phi(E_r, c_r, o_r),
\]
where $\Phi(\cdot)$ denotes the outcome-specific write-back rule. Its semantics can be expanded as
\[
E_{r+1} =
\begin{cases}
E_r \cup \{\text{accepted evidence}\}, & o_r = \mathrm{accepted}, \\
E_r \cup \{\text{negative evidence}\}, & o_r = \mathrm{rejected}, \\
E_r \cup \{\text{unstable evidence}\}, & o_r = \mathrm{unstable}, \\
E_r \cup \{\text{failure record}\}, & o_r = \mathrm{failed}.
\end{cases}
\]
This write-back map makes the outcome semantics explicit: accepted candidates produce confirmed positive evidence and model promotion, valid rejections produce confirmed negative evidence, unstable improvements produce weak positive evidence, and engineering failures write failure memory and repair traces.

The next-round research state is therefore written as
\[
\mathcal{S}_{r+1} = U(\mathcal{S}_r, c_r, o_r),
\]
where $U(\cdot)$ is the unified state-update map from Section~\ref{sec:method} and is expanded here through the two-layer updates of $M_{r+1}$ and $E_{r+1}$.

\begin{table*}[t]
\centering
\small
\caption{Round outcome semantics and state updates.}
\label{tab:s3_state_updates}
\begin{tabular}{llcccl}
\toprule
Outcome & Status & Evidence & Failure memory & Update best & Later effect \\
\midrule
Accepted candidate & confirmed positive & $\checkmark$ & $\times$ & $\checkmark$ & next-round starting point \\
Valid rejection & confirmed negative & $\checkmark$ & $\times$ & $\times$ & avoid the same mechanism \\
Unstable improvement & weak positive & $\checkmark$ & $\times$ & $\times$ & weak clue for later rounds \\
Engineering failure & no scientific conclusion & $\times$ & $\checkmark$ & $\times$ & constrain implementation paths \\
\bottomrule
\end{tabular}
\end{table*}

Table~\ref{tab:s3_state_updates} gives the unified interpretation of round outcomes, their write-back rule, and their effect on later planning. Accepted candidates update the current best. Valid rejections and unstable improvements enter the evidence state. Engineering failures update failure memory without producing scientific conclusions. Invalid protocol results are excluded from formal research evidence.

\paragraph{Next-Round Effect of Different Outcomes.}

The next Research Planner reads the full updated research state, including accepted candidates, negative results, unstable improvements, repair history, failure memory, and baseline diagnosis. This update rule turns historical execution into constraints and clues for later planning, which reduces repeated failure paths and unnecessary trial-and-error.

\paragraph{HTML Report Generation.}

When the research budget is exhausted or a termination condition is met, the system freezes the final current-best model, round history, canonical metrics, gate decisions, accepted and rejected evidence, and failure records. The Report Narrative Agent reads these frozen facts and generates a human-readable HTML report.

The Report Narrative Agent is responsible for narrative generation only. It cannot modify metrics, promotion decisions, model state, or evidence state. The final report is therefore a narrative output derived from frozen research facts.

The final output bundle includes the validated current-best model, reproducible source and configuration, token usage, and HTML report. The planner generates research intent, the builder forms candidate source states, the authority chain produces experimental facts, and the report agent turns frozen facts into an auditable result.

\section{Experimental Protocols and Reproducibility}
\label{app:protocols}

\subsection{Evaluation Protocol}

All baseline, ablation, and candidate evaluations follow the canonical forecasting benchmark path provided by the TFB repository.\footnote{\url{https://github.com/decisionintelligence/TFB}} EvoCast reuses the shared TFB evaluation protocol for agent-produced variants so that baseline establishment, baseline diagnosis, and later research rounds remain directly comparable under one metric surface.

The editable model pool also comes from the same repository family. EvoCast operates on forecasting models already exposed by the TFB codebase and treats them as the common source of baseline candidates and later architecture variants. This design keeps repository provenance, model construction, and metric computation inside one benchmark infrastructure.

\subsection{Agent Model and Pricing}

All planner, builder, and reporting calls use MiniMax-M3 as the unified foundation model backend. MiniMax-M3 provides up to 1M tokens of context with a guaranteed minimum of 512K tokens and serves as the single model family for the cognitive-agent layer in EvoCast.\footnote{\url{https://www.minimax.io/models/text/m3}}

\begin{table}[t]
\centering
\small
\caption{MiniMax-M3 API pricing used in our experiments.}
\label{tab:s4_minimax_pricing}
\begin{tabular}{ll}
\toprule
Item & Value \\
\midrule
Model & MiniMax-M3 \\
Service tier & standard \\
Context cap & $\leq 512$K input tokens \\
Uncached input price & 0.30 USD / M input tokens \\
Cache-hit input price & 0.06 USD / M input tokens \\
Output price & 1.20 USD / M output tokens \\
\bottomrule
\end{tabular}
\end{table}

Table~\ref{tab:s4_minimax_pricing} lists the MiniMax-M3 API prices used in our experiments under the actual setting: standard service tier with input context capped at 512K tokens. We report the official USD prices published by MiniMax and distinguish uncached input tokens from cache-hit input tokens.\footnote{\url{https://platform.minimax.io/docs/guides/pricing-paygo}}

\subsection{Budget Ledger and Alignment}

All agent workflows share a cap of at most 20 formal improvement rounds per task, with at most three repair attempts in each formal round. This rule aligns the formal improvement stage; it does not equate end-to-end resource consumption across workflows.

EvoCast performs task-specific baseline establishment and baseline mechanism diagnosis before formal rounds. AIDE uses a 10-round draft stage followed by at most 20 formal improvement rounds, while R\&D-Agent completes workflow initialization before formal research. Active seconds, LLM calls, total tokens, and API cost accumulate from workflow start through termination. Table~\ref{tab:rq2_resource_ledger_main} reports the per-task resource totals and valid-round ratios.

\begin{table*}[t]
\centering
\small
\caption{Formal-improvement budget and end-to-end resource-accounting boundary.}
\label{tab:budget_ledger}
\begin{tabular}{p{0.12\textwidth}p{0.20\textwidth}p{0.13\textwidth}p{0.12\textwidth}p{0.28\textwidth}}
\toprule
Workflow & Pre-formal stage & Formal cap & Repair cap & Resource-accounting scope \\
\midrule
AIDE & 10-round draft stage & 20 rounds & 3 & From draft-stage start through workflow termination. \\
R\&D-Agent & Workflow initialization & 20 rounds & 3 & From workflow start through termination. \\
EvoCast & Baseline establishment and mechanism diagnosis & 20 rounds & 3 & From baseline-establishment start through workflow termination. \\
\bottomrule
\end{tabular}
\end{table*}

\subsection{Supported Baseline Pool}

\begin{table*}[t]
\centering
\small
\caption{Supported baseline model pool grouped by architecture family.}
\label{tab:s5_baseline_pool}
\begin{tabular}{p{0.12\textwidth}cp{0.73\textwidth}}
\toprule
Category & Count & Models \\
\midrule
linear & 7 & CMoS~\citep{si2025cmos}, CrossLinear~\citep{zhou2025crosslinear}, DLinear~\citep{zeng2023are}, Linear~\citep{zeng2023are}, NLinear~\citep{zeng2023are}, SparseTSF~\citep{lin2024sparsetsf}, TimeBase~\citep{huang2025timebase} \\
transformer & 21 & Autoformer~\citep{wu2021autoformer}, Crossformer~\citep{zhang2023crossformer}, DTAF~\citep{lu2026dtaf}, DUET~\citep{qiu2025duet}, ETSformer~\citep{woo2022etsformer}, FEDformer~\citep{zhou2022fedformer}, Informer~\citep{zhou2021informer}, MultiPatchFormer~\citep{naghashi2025multipatchformer}, Nonstationary\_Transformer~\citep{liu2022nonstationary}, PAttn~\citep{tan2024arellms}, PDF~\citep{dai2024pdf}, PatchTST~\citep{nie2023patchtst}, Pathformer~\citep{chen2024pathformer}, Pyraformer~\citep{liu2022pyraformer}, Reformer~\citep{kitaev2020reformer}, TemporalFusionTransformer~\citep{lim2021tft}, TimeBridge~\citep{liu2025timebridge}, TimePerceiver~\citep{lee2025timeperceiver}, Transformer~\citep{vaswani2017attention}, Triformer~\citep{cirstea2022triformer}, iTransformer~\citep{liu2024itransformer} \\
mlp & 8 & FreTS~\citep{yi2023frets}, HDMixer~\citep{huang2024hdmixer}, LightTS~\citep{zhang2022lightts}, PatchMLP~\citep{tang2025patchmlp}, TSMixer~\citep{chen2023tsmixer}, TiDE~\citep{das2023tide}, TimeMixer~\citep{wang2024timemixer}, WPMixer~\citep{murad2025wpmixer} \\
cnn & 4 & MICN~\citep{wang2023micn}, ModernTCN~\citep{luo2024moderntcn}, SCINet~\citep{liu2022scinet}, TimesNet~\citep{wu2023timesnet} \\
gnn & 2 & MSGNet~\citep{zhang2024msgnet}, TimeFilter~\citep{hu2025timefilter} \\
others & 10 & Amplifier~\citep{fei2025amplifier}, FITS~\citep{xu2024fits}, FiLM~\citep{zhou2022film}, KANAD~\citep{zhou2025kanad}, Koopa~\citep{liu2023koopa}, MambaSimple~\citep{gu2023mamba}, SRSNet~\citep{wu2025srsnet}, SegRNN~\citep{lin2026segrnn}, TimeKAN~\citep{huang2025timekan}, xPatch~\citep{stitsyuk2025xpatch} \\
\bottomrule
\end{tabular}
\end{table*}

EvoCast establishes baselines from a supported pool of 52 forecasting architectures spanning linear, Transformer, MLP, CNN, GNN, and other model families. Table~\ref{tab:s5_baseline_pool} lists the full model pool grouped by architecture family.

\subsection{Execution Environment}

The current implementation is executed in the local base environment on an Ubuntu~22.04 system. The environment uses Python~3.12.7 with PyTorch~2.7.1+cu128, CUDA~12.8, NumPy~1.26.4, and pandas~2.2.2. Experiments run on one NVIDIA GeForce RTX~3080~Ti GPU with 12\,GB memory, 12 vCPU Intel(R) Xeon(R) Silver 4214R CPU @ 2.40\,GHz, and 90\,GB system memory.

\section{Additional Results and Analysis}
\label{app:results}

\subsection{RQ1 Additional Results}
\label{app:rq1}

RQ1 evaluates repository-level implementation capability under fixed predefined tasks. All compared methods receive the same initial repository snapshot, the same idea text, the same editable and forbidden regions, and the same canonical build/smoke entry. The comparison therefore measures whether a given architectural idea can be translated into faithful, protocol-compliant, and executable code under a shared validation pipeline.

The formal task suite contains 30 tasks built from 10 forecasting backbones and 3 idea-complexity levels per backbone. The 10 target backbones are DLinear, xPatch, FiLM, Crossformer, DUET, iTransformer, PatchTST, Amplifier, TimeFilter, and CrossLinear \citep{zeng2023are,stitsyuk2025xpatch,zhou2022film,zhang2023crossformer,qiu2025duet,liu2024itransformer,nie2023patchtst,fei2025amplifier,hu2025timefilter,zhou2025crosslinear}. All methods use MiniMax-M3 as the unified agent-driving model, and all compared agent methods use a fixed decoding temperature of 0.1.

The three idea-complexity levels are predefined task labels. For each backbone, the idea text is already given in advance, and the comparison focuses on whether an agent can faithfully translate that fixed idea into executable code under realistic deep-learning implementation constraints.

The compared methods are defined as follows.

\begin{itemize}
\item Direct Independent Edit: a one-shot direct repository edit from the given idea text, without multi-step planning, bounded repair, or agent-level orchestration.
\item mini-SWE-agent~\citep{yang2024sweagent}: a lightweight software-engineering agent that iteratively edits the repository with tool use and execution feedback.
\item AutoResearch~\citep{kumar2026autoresearch}: an execution-grounded research workflow that organizes the task into iterative proposal, implementation, and validation steps.
\item AIDE~(2025) \citep{jiang2025aide}: a search-based ML engineering agent that explores and evaluates candidate implementation trajectories.
\item R\&D-Agent~\citep{yang2025rdagent}: a dual-role research-and-development workflow in which proposal generation and code realization are handled by separate agent functions.
\end{itemize}

Together, these baselines cover one-shot editing, lightweight SWE agents, search-based engineering, execution-grounded research workflows, and split-role research/development pipelines.

\begin{table*}[t]
\centering
\small
\caption{The three predefined idea-complexity levels used in RQ1.}
\label{tab:rq1_idea_levels}
\begin{tabular}{p{0.16\textwidth}p{0.30\textwidth}p{0.24\textwidth}p{0.20\textwidth}}
\toprule
Level & Definition & Typical forms & Main capability under test \\
\midrule
Level~1: Local Modification & Modify an existing module locally without changing the surrounding execution structure. & Replace an activation; add a residual gate; adjust the normalization position; insert a lightweight channel mixer. & Small and precise local code edits inside an existing deep-learning block. \\
Level~2: Cross-Module Integration & Understand multiple source regions and connect existing modules through a new live path. & Add adaptive fusion between decomposition and the forecasting head; inject a frequency branch into the backbone; align multi-scale paths and then fuse them. & Cross-file and cross-module implementation ability, including interface matching and shape alignment. \\
Level~3: Novel Module Insertion & Add a new module or reorganize the main execution path so that the new mechanism participates in prediction. & Add a context-conditioned spectral gate; add a horizon-adaptive branch; add a variable interaction router. & New-module implementation and main-path restructuring under executable repository constraints. \\
\bottomrule
\end{tabular}
\end{table*}

Table~\ref{tab:rq1_main_results} summarizes the method-level RQ1 results on all 30 tasks. The metrics are defined as follows.

\begin{itemize}
\item Success: the proportion of tasks that pass the canonical build/smoke pipeline.
\item Fidelity: the proportion of evaluated implementations that faithfully realize the given idea and avoid semantically mismatched runnable code.
\item Repair: repair success rate computed over repair-eligible tasks only, i.e., repair-success count divided by repair-sample count.
\item Avg. Token: average total token usage per task, including uncached input, cache-hit input, and output tokens.
\item Avg. Cost (USD): average API cost per task under the MiniMax-M3 pricing scheme.
\item Avg. Time (s): average wall-clock runtime per task from execution start to final task outcome.
\end{itemize}

EvoCast achieves the highest overall success rate at 96.7\%, followed by mini-SWE-agent at 90.0\%, AutoResearch at 86.7\%, and AIDE (2025) at 83.3\%. Direct Independent Edit has the lowest average token, cost, and wall-clock time, but also the lowest success rate. EvoCast combines the highest executable success with the second-lowest average token, cost, and time, which indicates strong execution efficiency under the shared protocol.

\begin{figure}[t]
\centering
\includegraphics[width=\columnwidth]{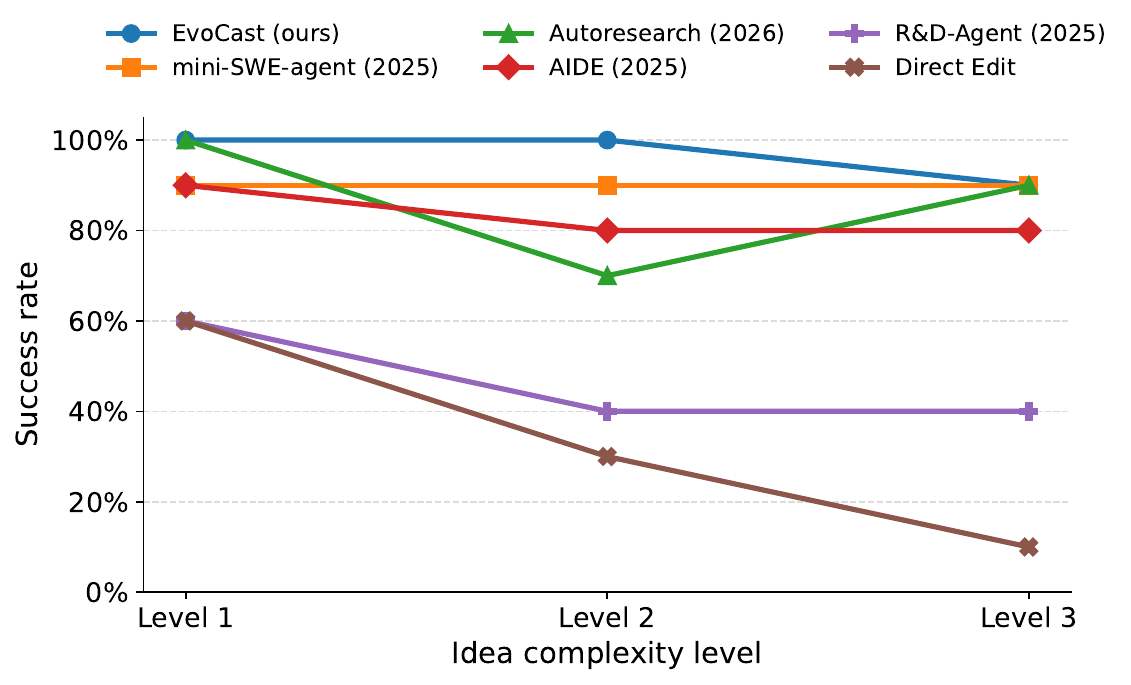}
\caption{Success-rate trends across idea-complexity levels.}
\Description{Line chart of implementation success rates for several agent methods across low, medium, and high idea-complexity levels. EvoCast maintains the strongest success trend as complexity increases.}
\label{fig:rq1_level_trends}
\end{figure}

Figure~\ref{fig:rq1_level_trends} shows how implementation reliability changes as the required modification depth increases from local edits to cross-module integration and then to novel-module insertion. The main pattern is that methods differ sharply in how gracefully they degrade under this increase in implementation complexity. EvoCast remains stable from Level~1 to Level~2 and only shows a limited drop at Level~3, which indicates that its execution framework continues to hold when the task begins to require new modules and main-path restructuring. mini-SWE-agent is also relatively stable across levels, while Direct Independent Edit deteriorates rapidly as soon as the task moves beyond local modification. This contrast shows that RQ1 evaluates more than basic code generation ability: the harder levels expose whether a method can preserve executable, semantically aligned implementation when the edit must cross files, alter live computation paths, or introduce new architectural components.

The backbone-level results show that the main difference across methods includes both overall success frequency and the breadth of transfer across heterogeneous forecasting repositories. Direct Independent Edit has the widest failure spread: it completely fails on Amplifier, CrossLinear, Crossformer, and iTransformer, and remains unstable on PatchTST, TimeFilter, and xPatch. This pattern indicates that one-shot editing breaks down once the target model requires deeper coordination across existing modules, execution paths, and repository-specific implementation conventions.

The agent baselines are more stable than direct editing, but their failures remain concentrated on a small set of harder backbones. AIDE~(2025) \citep{jiang2025aide} drops mainly on Crossformer, iTransformer, and DLinear. AutoResearch~\citep{kumar2026autoresearch} is weakest on iTransformer, Crossformer, and xPatch. mini-SWE-agent~\citep{yang2024sweagent} stays strong on most backbones but still collapses on xPatch. R\&D-Agent~\citep{yang2025rdagent} shows the broadest instability among the agent baselines, with failures distributed across FiLM, Amplifier, DUET, PatchTST, TimeFilter, and iTransformer. These concentrated failure clusters suggest that average success rate alone is not sufficient: methods with similar overall scores can still differ substantially in how robustly they handle different repository structures.

Against this background, EvoCast is distinguished by the narrowest failure profile and the most uniform behavior across backbones. Its advantage therefore comes from cross-backbone consistency and does not depend on isolated wins on a few easier targets. This is the main reason the backbone-level analysis matters for RQ1: the benchmark evaluates whether a method can reliably implement ideas in realistic, heterogeneous deep-learning repositories across the full task set.

\subsection{RQ2 Additional Results}
\label{app:rq2}

RQ2 evaluates whether the same fixed EvoCast workflow can develop stronger task-specific architectures for real forecasting tasks under controlled agent-side cost. This section examines final error, final solution type, and how effectively a fixed budget is converted into measurable research progress. All error metrics are normalized, and lower values are better.

RQ2 contains three fixed real forecasting tasks that cover MM, SS, and MS settings. For each task, the dataset, lookback, horizon, forecasting mode, evaluation metric, training budget, and research-round budget are fixed before the autonomous run begins. All baselines, ablations, candidates, and final replay runs use the canonical benchmark path of the TFB repository \citep{qiu2024tfb}, so evaluator logic, task definition, and metric surface remain fixed.

\begin{table*}[t]
\centering
\small
\caption{RQ2 tasks and dataset characteristics.}
\label{tab:rq2_tasks}
\begin{tabular}{p{0.17\textwidth}c c c L{0.15\textwidth} p{0.30\textwidth}}
\toprule
Dataset & Mode & Rows & Setting & Variables & Task characteristics \\
\midrule
Air Quality \citep{devito2008airquality} & MM & 9,357 & 168 $\rightarrow$ 24 & 12 air-quality, sensor-response, and weather variables & Multivariate air-quality forecasting with cross-variable dependency, seasonal structure, and trend variation. \\
Melbourne Pedestrian T33 \citep{godahewa2020melbourne} & SS & 47,315 & 168 $\rightarrow$ 24 & Univariate hourly pedestrian count & Single-series pedestrian-flow forecasting with strong daily and weekly periodicity and long-term level change. \\
Steel Industry \citep{sathishkumar2021steel} & MS & 35,040 & 720 $\rightarrow$ 120 & 9 input variables and target Usage\_kWh & Multivariate-to-univariate industrial power forecasting with long lookback, long horizon, regime transition, seasonal structure, and distribution drift. \\
\bottomrule
\end{tabular}
\end{table*}

The three RQ2 tasks are all built on publicly available datasets and cover clearly different real forecasting scenarios. Air Quality \citep{devito2008airquality} comes from hourly pollutant concentrations and gas-sensor responses collected at a road-level monitoring station in an Italian city, which forms a multivariate air-quality forecasting task. Melbourne Pedestrian T33 \citep{godahewa2020melbourne} comes from hourly pedestrian counts recorded by the City of Melbourne pedestrian counting system, and we use the public T33 series as a single-series forecasting task. Steel Industry \citep{sathishkumar2021steel} comes from electricity-consumption records collected in a smart small-scale steel industry in South Korea, and we use it as a multivariate-to-univariate industrial energy forecasting task with Usage\_kWh as the prediction target.
Table~\ref{tab:rq2_tasks} summarizes the task settings and core characteristics of the three public datasets.

All tasks follow the same experimental protocol: the data are split by chronological order with a 7:1:2 ratio for training, validation, and testing; no split is shuffled; sample construction uses rolling windows with stride 1; normalization statistics are computed on the training set only and then reused for validation and testing; and MSE and MAE are computed in that normalized space.

For general-purpose time-series forecasting models, we use a shared base training budget with batch size 32, a maximum of 10 epochs, and early-stopping patience 3. Beyond these shared training constraints, model-specific hyperparameters are tuned within predefined reasonable ranges to obtain comparable task-level performance. For agent-based methods, we keep the training hyperparameters specified by the agent workflow for its final selected program and then reevaluate that program under the same formal forecasting protocol. This setting lets strong forecasting models and agent workflows follow their natural usage patterns while sharing the same data split, prediction windows, normalization rule, and evaluation metrics.

The formal comparison includes only artifacts that satisfy the forecasting-time information constraint and the shared evaluation protocol. This setting keeps the focus on final deliverable forecasting artifacts under one fixed evaluator.

All cognitive-agent calls use MiniMax-M3 with thinking mode set to auto. Temperature follows each method's own program setting; when a method does not explicitly pass a temperature value, no additional temperature override is injected at runtime. EvoCast uses the formal run with seed 2027; when needed, archived seed-evaluation results are used as stability references. The metrics used in this section are defined as follows:

\begin{itemize}
\item MSE: normalized mean squared error on the formal rolling test windows.
\item MAE: normalized mean absolute error on the formal rolling test windows.
\item Active Sec.: wall-clock seconds consumed by the complete agent workflow from start to termination, including planning, coding, execution, repair, evaluation, and state update.
\item LLM Calls: total number of model calls made during the complete agent run.
\item Total Tokens: cumulative input and output token usage.
\item Cost USD: total API cost under the actual pricing setting.
\end{itemize}

The compared methods cover both strong forecasting models and autonomous agent workflows. Together, they span foundation-model forecasting, multimodal forecasting, machine-learning engineering agents, and research-oriented agent workflows. The compared methods are defined as follows:

\begin{itemize}
\item Sundial~(2025) is a generative time-series foundation model that emphasizes scalable pretraining and zero-shot forecasting.
\item Toto-2.0-2.5B~(2026) is a large-scale time-series foundation model that emphasizes forecasting performance gains with model scaling.
\item Time-VLM~(2025) \citep{zhong2025timevlm} is a multimodal forecasting model that enhances time-series representations with visual and textual signals.
\item TimeCMA~(2025) \citep{liu2025timecma} is an LLM-empowered multivariate forecasting model that emphasizes multivariate prediction under cross-modality alignment.
\item AIDE~(2025) \citep{jiang2025aide} is a machine-learning engineering agent that models iterative trial-and-error as tree search in code space.
\item R\&D-Agent~(2025) \citep{yang2025rdagent} is a dual-agent research-and-development workflow in which a Researcher proposes directions and a Developer executes implementation with feedback-driven iteration.
\item EvoCast is the proposed fixed-protocol autonomous architecture-development workflow with task-specific baseline diagnosis, evidence-grounded planning, bounded repository-level implementation, canonical evaluation, and program-controlled promotion.
\end{itemize}

The formal-improvement protocol caps each agent task at 20 research rounds, with at most three repair attempts per round. This cap aligns the formal stage only; Table~\ref{tab:budget_ledger} records the distinct pre-formal stages and end-to-end accounting boundaries. AIDE uses a 10-round draft stage before its formal improve stage, while EvoCast establishes and diagnoses a task-specific baseline before formal research. EvoCast selects FreTS for Air Quality \citep{yi2023frets}, PatchMLP for Melbourne Pedestrian T33 \citep{tang2025patchmlp}, and PatchMLP for Steel Industry \citep{tang2025patchmlp} as the starting baselines.

Table~\ref{tab:rq2_overall_results_main} reports the final formal forecasting results on the three tasks. For agent methods, each value comes from the final selected program under the unified evaluator with fixed seed 2027. EvoCast reduces the selected-baseline error on all three tasks. It achieves the lowest MSE on all three tasks and the lowest MAE on two of them; AIDE (2025) attains the lowest MAE on Melbourne Pedestrian T33. These results show that the fixed EvoCast loop can move the search toward stronger task-specific architectures without changing the evaluator or the task protocol.

Table~\ref{tab:rq2_overall_results_main} also shows where different workflows finally converge. EvoCast stays on repository-native forecasting architecture evolution and delivers a stronger final artifact through continued live-path development. The competing agent workflows more often converge to broader and more conservative solution types. The comparison therefore reflects final deliverable architecture development under one shared protocol.

\begin{table*}[t]
\centering
\small
\caption{EvoCast versus general-purpose SOTA forecasting models on RQ2 tasks. Best values are \textbf{bolded} and second-best values are \underline{underlined}.}
\label{tab:rq2_sota_forecasting_models}
\begin{tabular}{p{0.25\textwidth}C{0.09\textwidth}C{0.09\textwidth}C{0.09\textwidth}C{0.09\textwidth}C{0.09\textwidth}C{0.09\textwidth}}
\toprule
Method & \multicolumn{2}{c}{Air Quality} & \multicolumn{2}{c}{Melbourne Pedestrian T33} & \multicolumn{2}{c}{Steel Industry} \\
\cmidrule(lr){2-3}\cmidrule(lr){4-5}\cmidrule(lr){6-7}
 & MSE & MAE & MSE & MAE & MSE & MAE \\
\midrule
DLinear (2023) & 0.5160 & 0.5344 & 0.2655 & 0.3149 & 0.3268 & 0.3722 \\
NLinear (2023) & 0.6004 & 0.5692 & 0.3188 & 0.3737 & 0.3476 & 0.3975 \\
FreTS (2023) & \underline{0.4665} & 0.5004 & \underline{0.1844} & 0.2446 & 0.3285 & 0.3847 \\
FITS (2024) & 0.6672 & 0.6151 & 0.4179 & 0.4360 & 0.3572 & 0.3966 \\
HDMixer (2024) & 0.4753 & \underline{0.4950} & 0.2271 & 0.2663 & 0.3318 & \underline{0.3395} \\
Amplifier (2025) & 0.4709 & 0.4956 & 0.2049 & 0.2636 & 0.3188 & 0.3674 \\
DUET (2025) & 0.9429 & 0.7264 & 0.5783 & 0.5610 & 0.4882 & 0.4631 \\
PatchMLP (2025) & 0.4850 & 0.5064 & 0.1852 & \underline{0.2410} & \underline{0.3057} & 0.3733 \\
TimeBase (2025) & 0.7047 & 0.6477 & 0.4355 & 0.4041 & 0.7762 & 0.7442 \\
TimeBridge (2025) & 0.6445 & 0.6037 & 0.2483 & 0.2990 & 0.3595 & 0.4139 \\
TimeKAN (2025) & 0.4732 & 0.4973 & 0.2212 & 0.2725 & 0.3311 & 0.3690 \\
DTAF (2026) & 0.4994 & 0.5137 & 0.2532 & 0.2902 & 0.3173 & 0.3450 \\
SegRNN (2026) & 0.5220 & 0.5269 & 0.1955 & 0.2491 & 0.4216 & 0.4513 \\
EvoCast & \textbf{0.4570} & \textbf{0.4909} & \textbf{0.1628} & \textbf{0.2242} & \textbf{0.2792} & \textbf{0.3388} \\
\bottomrule
\end{tabular}
\end{table*}

Table~\ref{tab:rq2_sota_forecasting_models} compares the task-specific architectures discovered by EvoCast with general-purpose SOTA time-series forecasting models. Each forecasting model is directly trained and evaluated on each RQ2 task under the same TFB protocol. EvoCast first runs the autonomous architecture-development workflow and then evaluates the final discovered architecture under the same protocol. This comparison evaluates whether fixed-workflow autonomous architecture development can produce stronger task-specific forecasting models than direct use of strong general-purpose forecasting architectures.

EvoCast achieves lower MSE and MAE than all listed general-purpose forecasting models on all three tasks. This result shows that the EvoCast-developed architectures improve the selected starting backbones and form stronger task-specific solutions than the shared SOTA forecasting-model set under the same evaluation protocol.

Table~\ref{tab:rq2_resource_ledger_main} quantifies complete agent-run resource consumption, including LLM calls, tokens, API cost, and active runtime. Foundation forecasting models only perform inference, so this table focuses on AIDE (2025), R\&D-Agent, and EvoCast. The table measures agent-side resource consumption, and its interpretation should be combined with later evidence on valid rounds, trajectory stability, and final artifact type. EvoCast stays far below R\&D-Agent in token usage and API cost on all three tasks. Relative to AIDE (2025), EvoCast uses more tokens and API cost on Air Quality and Steel Industry, and more tokens but lower API cost on Melbourne T33; the relative wall-clock time varies by task. Table~\ref{tab:rq2_overall_results_main} reports the corresponding final forecasting quality. These resource totals are interpreted alongside round validity and final forecasting quality under the shared formal-round cap.

Table~\ref{tab:rq2_resource_ledger_main} quantifies how often each agent workflow turns a formal round into a measurable result. EvoCast keeps high round validity on all three tasks. Higher round validity directly increases the amount of formal evidence available to later rounds and supports stronger accumulation in the later architecture and trajectory analysis. The resource consumption in Table~\ref{tab:rq2_resource_ledger_main} therefore separates into two different outcomes: one part of the budget is spent on invalid rounds, more volatile attempts, and weaker terminal convergence, while another part is continuously deposited into reusable formal evidence and stronger final architectures.

\begin{figure}[t]
\centering
\includegraphics[width=\columnwidth]{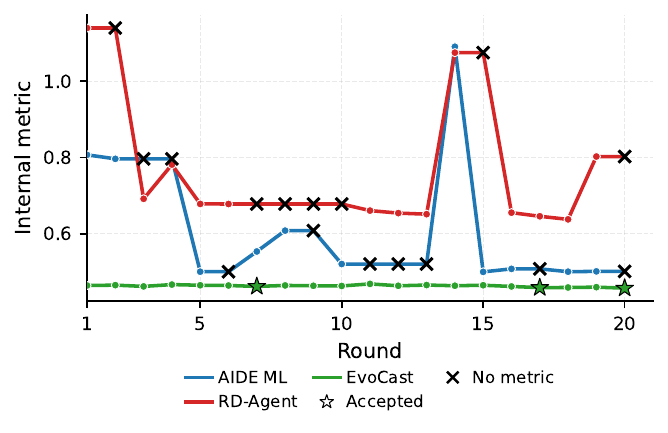}
\caption{Round-level internal trajectories of agent workflows on Air Quality. EvoCast stays in a lower and more stable error band, while AIDE (2025) and R\&D-Agent show larger fluctuations and more rounds without usable metrics.}
\Description{Round-by-round Air Quality error trajectories for EvoCast, AIDE, and R and D-Agent. EvoCast remains in a lower and more stable error band, while the other methods fluctuate and include missing usable metrics.}
\label{fig:rq2_agent_air_trajectories}
\end{figure}

Figure~\ref{fig:rq2_agent_air_trajectories} further distinguishes the competing agent workflows through process stability and final structural landing point. On Air Quality, EvoCast stays in a lower error band with smaller fluctuation, and its accepted rounds are concentrated near the best region. AIDE (2025) and R\&D-Agent fluctuate more strongly and lose more rounds to non-usable outcomes. The same fixed budget therefore produces sharply different amounts of formal evidence and leads to different convergence directions.

The final selected artifacts expose a second difference. On Air Quality, AIDE (2025) converges to a linear Ridge feature-engineering solution, and R\&D-Agent converges to a GRU/LSTM ensemble. These solutions are runnable, but they land on more conservative structural families and drift away from repository-native forecasting architecture evolution. EvoCast keeps the search on the live path of the selected backbone, so the final artifact remains a task-specific forecasting architecture developed inside the target repository. This difference shows a clear split in final artifact type under the same formal-round cap: EvoCast continues to deliver deep forecasting architectures with newly accumulated active mechanisms, while the competing agent workflows more often converge to broader and more conservative solution types.

\begin{figure*}[t]
\centering
\includegraphics[width=\textwidth]{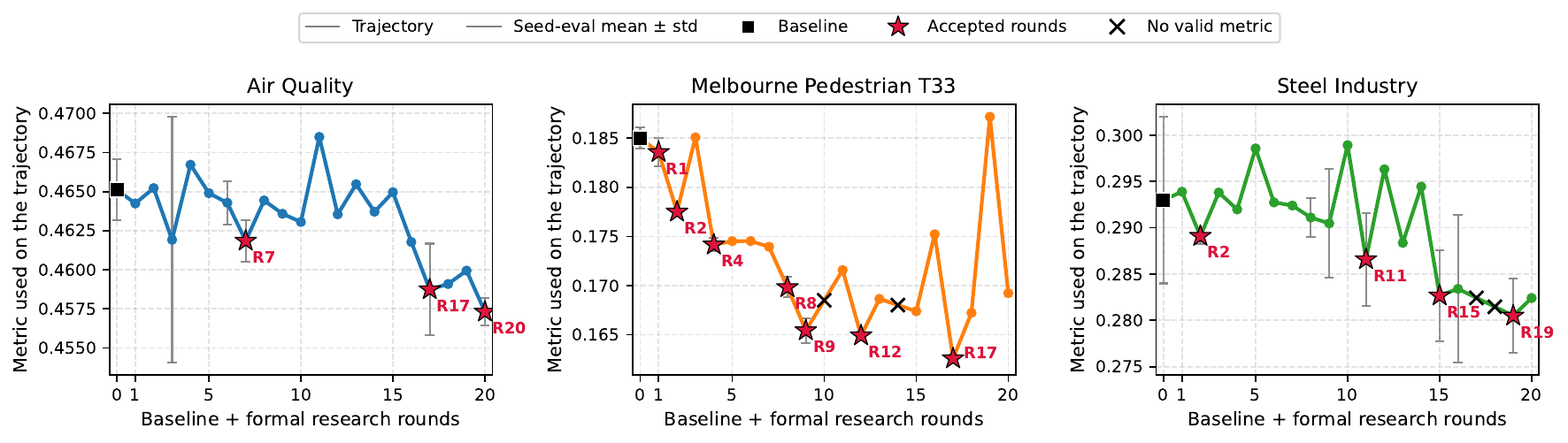}
\caption{Baseline and round-level trajectories on the three RQ2 real forecasting tasks. Red stars denote accepted rounds, and error bars show available seed-level stability information.}
\Description{Three task-level plots showing selected baselines and round-level forecast errors. Red stars mark accepted research rounds and error bars indicate available seed-level stability information.}
\label{fig:rq2_iteration_curves}
\end{figure*}

Figure~\ref{fig:rq2_task_adaptive_architectures} further answers whether the same fixed workflow preserves different active structures across tasks and whether those structures are formed through multi-round accumulation. The figure shows the final task-adaptive live-path architectures selected by EvoCast. The three runs follow the same protocol, but the retained active modules differ substantially across MM, SS, and MS tasks. Air Quality mainly accumulates cross-channel trend anchoring and horizon-shape refinement. Melbourne Pedestrian T33 accumulates periodic memory, calendar-conditioned retrieval, and horizon-wise output shaping. Steel Industry accumulates recency-aware refinement, dual-path projection, and output-side correction. This figure brings the earlier cost and trajectory analysis down to the final artifact level: the budget consumed by EvoCast is ultimately deposited into task-specific active modules that remain on the original backbone live path.

Figure~\ref{fig:rq2_iteration_curves} places the baseline, formal research rounds, and accepted rounds into a shared coordinate system. Air Quality improves more gradually, with key gains appearing in the middle and later rounds. Melbourne Pedestrian T33 obtains clear gains earlier and then converges through subsequent refinement. Steel Industry shows a more stable evolution path. The final structures therefore come from multi-round research accumulation and correspond to a traceable path of architecture evolution.

On Air Quality, EvoCast starts from FreTS and forms a frequency-domain architecture with cross-channel trend anchoring, cross-horizon interaction, and channel-conditioned horizon-axis warping. The task shows strong cross-variable dependency, pronounced transition behavior, and a clear trend component, so the final architecture concentrates on cross-channel trend stabilization and horizon-shape correction. The resulting model extends the original frequency backbone into a task-specific structure that jointly corrects channel relations and forecast-trajectory shape.

On Melbourne Pedestrian T33, EvoCast starts from PatchMLP and forms an accumulated architecture centered on periodic prior injection, representation modulation, and output-distribution correction. The task has strong daily and weekly periodicity together with visible regime transition, so the live path retains periodic calendar memory, calendar-conditioned retrieval, lag-shift similarity refinement, period-aware decomposition gating, lookback-value token conditioning, horizon-step target-distribution mixture, and cross-window regime refinement. The final architecture therefore expresses a complete periodic-awareness chain with coordinated modules.

On Steel Industry, EvoCast again starts from PatchMLP and converges to a structure oriented toward long-lookback and long-horizon forecasting. The task combines strong transition behavior, seasonality, and trend, which moves the search toward output organization, horizon-dependent bias, and recency-aware correction. The final live path retains dual-path trend/seasonal projection, horizon-bias pre-projection, quantile-indexed output-distribution correction, and channel-sparse recency mixture. The resulting structure integrates long-range dependency handling, output-distribution correction, and sparse recent-context information on one live path.

RQ2 compares final deliverable artifacts under one unified formal protocol. For every agent workflow, the final selected program is reevaluated by the same external evaluator used for the baselines and strong forecasting models. Intermediate artifacts that depend on information unavailable at forecasting time do not enter the formal comparison. The reported results therefore focus on forecasting artifacts that satisfy the shared TFB evaluation protocol.

\subsection{RQ3 Additional Results}
\label{app:rq3}

RQ3 examines how EvoCast's system-level components affect autonomous architecture-development reliability, final quality, and agent-side cost. All ablations are defined against the full EvoCast workflow under the same forecasting task, the same baseline, and the same research budget. The goal is to isolate how each component contributes to the full loop from evidence construction to research planning, bounded execution, and iterative state update.

The task is fixed to Melbourne Pedestrian T33 with PatchMLP as the baseline, lookback 168, horizon 24, and optimization metric MSE. Each run contains 20 formal research rounds, and the repair budget is 3. A valid round is a formal research round that produces a measurable MSE or MAE. The full EvoCast system contains dataset characterization, baseline diagnosis, evidence-grounded planning, authority-controlled execution, and iterative evidence update. RQ3 removes one component at a time while keeping the remaining pipeline fixed.

Full system setting. The full system executes the complete EvoCast loop and serves as the reference point for all ablations. It preserves task evidence construction, executed baseline diagnosis, evidence-grounded planning, bounded repository-level implementation, and iterative state write-back across rounds.

A1: Dataset characterization removed. This variant removes dataset-profile input from the planner while keeping baseline diagnosis, formal planning, authority-controlled execution, and iterative state update unchanged. The removed information is the explicit task evidence that describes data structure, temporal pattern, and forecasting difficulty. A1 therefore tests how much direction quality depends on direct task evidence rather than only on baseline behavior and later-round execution records.

A2: Baseline diagnosis removed. This variant keeps dataset characterization and the later formal research loop, but it removes baseline mechanism ablation before research rounds begin. The planner still sees task evidence, yet it no longer receives executed evidence about which baseline mechanisms are already effective, weak, or redundant. A2 therefore tests whether mechanism-level starting-point judgment is necessary for targeted module selection and efficient budget allocation.

A3: Evidence-grounded planning removed. This variant keeps dataset characterization, baseline diagnosis, execution, and evaluation, but it removes the planner's structured evidence inputs, including dataset profile, baseline diagnosis, research context, round history, and recent status. Candidate construction can still proceed, but direction selection no longer uses the complete accumulated evidence bundle. A3 therefore tests whether continuous executability alone is sufficient, or whether high-quality convergence requires fully evidence-grounded direction selection.

A4: Authority-controlled execution removed. This variant keeps research-direction generation and multi-round execution, but it removes repository write-boundary control in the candidate workspace. The planner still produces mechanism hypotheses, yet candidate implementation is no longer restricted to source regions directly tied to the current hypothesis. A4 therefore tests whether execution-boundary control is necessary to keep repository edits focused, maintain budget discipline, and preserve reliable formal comparison across rounds.

A5: Iterative evidence update removed. This variant keeps the initial task evidence, baseline diagnosis, planning, and bounded execution, but it freezes the historical evidence visible to the planner so later-round results are no longer written back. The current best model can still be retained, but failure records, marginal improvements, repair history, and validated mechanisms no longer enter later planning. A5 therefore tests whether persistent evidence accumulation is necessary for sustained multi-round progress.

Table~\ref{tab:rq3_ablation_results_main} reports final forecasting quality, valid rounds, and agent-side cost. Final MSE uses the final retained artifact and reports the three-seed mean over seeds 2026, 2027, and 2028; when an ablation fails to produce a promotable candidate, the retained baseline value is reported. Valid rounds counts the number of formal research rounds that produce a measurable forecasting metric. Time, Tokens, and Cost come from the HTML round-audit records.

Table~\ref{tab:rq3_ablation_results_main} gives a direct result: the full EvoCast loop achieves the lowest MSE while keeping runtime, token usage, and API cost controlled. Every ablation weakens this balance, and each one exposes a different kind of system distortion. The full-system advantage therefore comes from the complete closed loop.

A1, A2, and A3 reveal the source of direction quality. Removing dataset characterization drops the final result to 0.173547, which shows that task evidence directly affects the structural direction of later rounds. Removing baseline diagnosis yields 0.173473 while increasing token usage and cost to 2,794,140 and 1.4694 USD. Baseline diagnosis therefore provides the judgment of which mechanisms in the current backbone are already effective and which regions deserve targeted modification. Removing evidence-grounded planning yields 0.173250. This variant still produces measurable results in 20/20 formal rounds, but the final model remains clearly weaker than the full system. The result shows that continuous executability alone does not move the search toward the highest-quality structural region; the evidence bundle determines the quality of direction selection.

A4 and A5 reveal the source of process reliability. Removing authority-controlled execution causes the research process to lose focus quickly. Once repository write boundaries are removed, the LLM frequently inspects external files unrelated to the current candidate, edits source regions outside the main modification path, and uses broad trial-and-error to compensate for uncertainty about repository structure. These behaviors sharply increase runtime, token usage, and cost, reaching 10.80 hours, 25,321,252 tokens, and 7.8792 USD. The retained final artifact still stays at the baseline level of 0.185001. This result shows that authority-controlled execution maintains research focus and budget efficiency by restricting candidate implementation to source regions directly tied to the current mechanism hypothesis. Removing iterative evidence update degrades the final result to 0.178489 with 15/20 valid rounds. This ablation keeps the initial evidence and the current-best model, but the planner can no longer absorb later failure records, marginal improvements, or validated mechanisms. The process therefore plateaus earlier. Together, A4 and A5 show that execution-boundary control and persistent evidence write-back keep multi-round research focused, comparable, and accumulative.

Figure~\ref{fig:rq3_event_only} provides dynamic evidence for the same conclusion. The plot shows the cumulative best trajectories of Full, A1, A2, A3, and A5. The full EvoCast curve continues to descend and still refreshes the best value in the later half of the run. A5 enters a plateau earlier. A1, A2, and A3 all obtain multiple improvements, but each stops at a clearly higher level than the full system. This trajectory view shows that the full system improves both final error and sustained research progress.

\section{Code and Usage}
\label{app:code}

The public code repository is available at \url{https://github.com/18e0-x/EvoCast}.

\begin{figure*}[t]
\centering
\includegraphics[width=\textwidth]{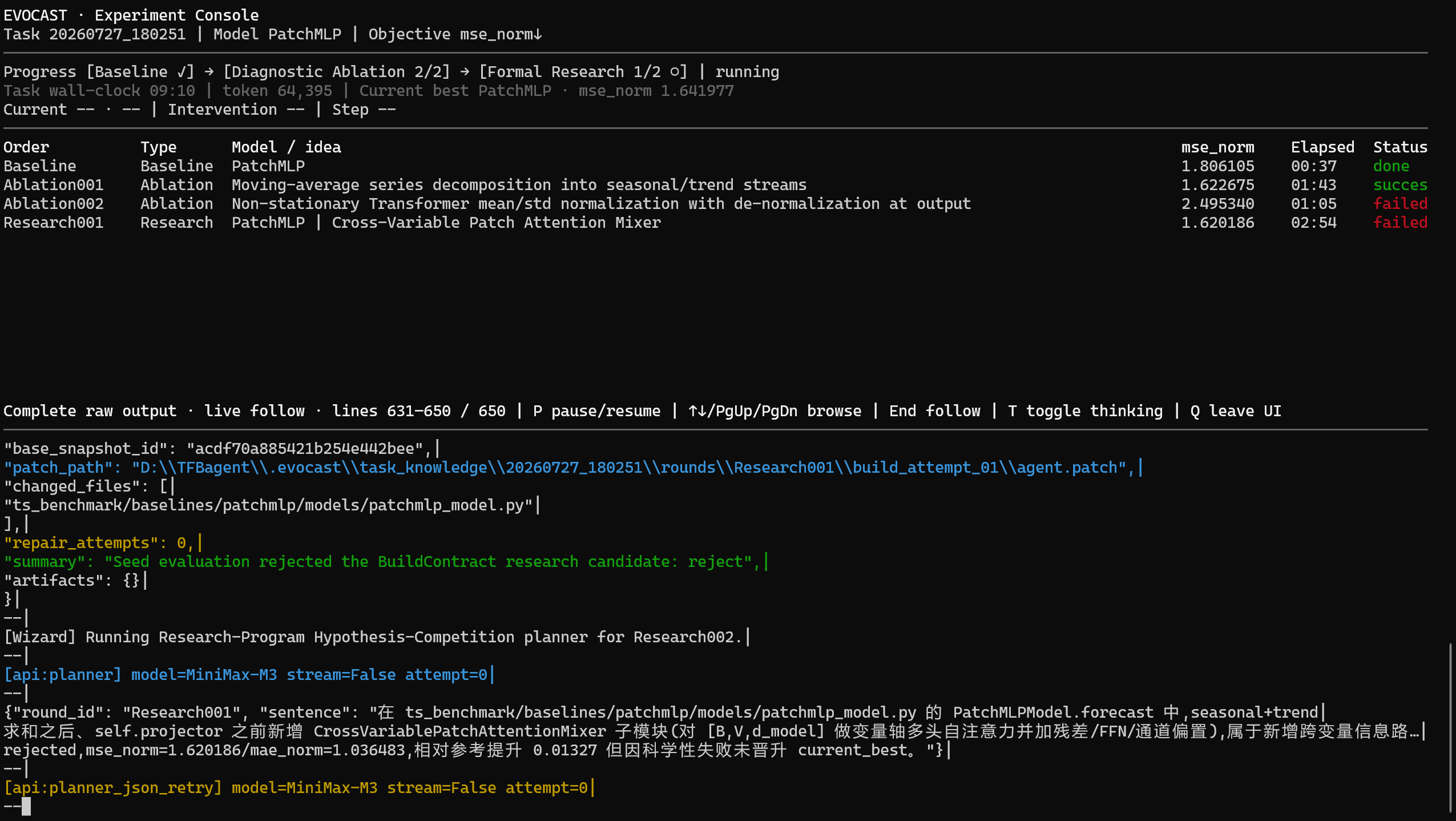}
\caption{EvoCast terminal experiment console. The interface unifies live progress tracking, round-level status, metrics, elapsed time, token-aware execution context, and raw-log inspection in one view.}
\Description{Screenshot of the EvoCast terminal console. It shows task and stage status, the current best result, baseline and research-round summaries, elapsed time, token accounting, and live raw logs.}
\label{fig:terminal_console}
\end{figure*}

The unified entry, \texttt{python -m evocast}, opens a wizard that collects the task definition, budget, baseline strategy, and research start conditions. It then connects task configuration, baseline establishment, dataset characterization, baseline diagnosis, formal research, and finalization under one fixed protocol. Environment preparation and usage details are provided in the public repository's README.

The full system uses the default configuration, and A1--A5 use the same entry and task lifecycle. A1 uses \texttt{--dataset-\allowbreak diagnosis-\allowbreak mode skip} to remove dataset characterization. A2 uses \texttt{--max-\allowbreak ablation-\allowbreak targets 0} to skip baseline mechanism ablations. A3 uses \texttt{--agent-ablation A3} to remove planner evidence injection. A4 uses \texttt{--agent-ablation A4} to open the editable authority scope. A5 uses \texttt{--agent-ablation A5} to freeze iterative evidence update. The variants share the execution and finalization interfaces used by the main experiments.

Build mode, selectable in the wizard, supports fast end-to-end reliability checks. It retains task initialization, baseline selection, characterization, diagnosis, formal research, state updates, gate flow, and HTML reporting, while reducing the training, validation, and testing budget. It supports checks of task compilation, protocol execution, edit boundaries, runtime feedback, and report generation before longer formal runs.

Figure~\ref{fig:terminal_console} shows the online inspection interface. The upper area displays the task id, active model, objective, stage progress, current best result, and cumulative wall-clock and token usage. The middle table summarizes baseline, ablation, and research rounds with their status, metrics, and elapsed time. The lower area preserves raw execution output, including failure reasons and planner responses. Live follow, pause/resume, and log browsing support monitoring during execution.

Offline inspection uses the HTML review report and final model pack. The report preserves the research process and conclusions, while the model pack exports the retained model, configuration, and metric records for retrospective verification. Accompanying reproducibility materials include the data-preparation script, processed task datasets, and EvoCast system, exposing the preprocessing path and task inputs and supporting workflow execution, ablation inspection, and result export.

%% file: references.bib
@article{qiu2024tfb,
  title   = {TFB: Towards Comprehensive and Fair Benchmarking of Time Series Forecasting Methods},
  author  = {Xiangfei Qiu and Jilin Hu and Lekui Zhou and Xingjian Wu and Junyang Du and Buang Zhang and Chenjuan Guo and Aoying Zhou and Christian S. Jensen and Zhenli Sheng and Bin Yang},
  journal = {Proc. {VLDB} Endow.},
  volume  = {17},
  number  = {9},
  pages   = {2363--2377},
  year    = {2024}
}

@article{devito2008airquality,
  title   = {On Field Calibration of an Electronic Nose for Benzene Estimation in an Urban Pollution Monitoring Scenario},
  author  = {Simone De Vito and Ettore Massera and Mauro Piga and Luca Martinotto and Girolamo Di Francia},
  journal = {Sensors and Actuators B: Chemical},
  volume  = {129},
  number  = {2},
  pages   = {750--757},
  year    = {2008},
  doi     = {10.1016/j.snb.2007.09.060}
}

@misc{godahewa2020melbourne,
  title        = {Melbourne Pedestrian Counts Dataset},
  author       = {Rakshitha Godahewa and Christoph Bergmeir and Geoffrey I. Webb and Rob J. Hyndman and Pablo Montero-Manso},
  year         = {2020},
  doi          = {10.5281/zenodo.4656626},
  howpublished = {Zenodo},
  url          = {https://doi.org/10.5281/zenodo.4656626}
}

@article{sathishkumar2021steel,
  title   = {Efficient Energy Consumption Prediction Model for a Data Analytic-Enabled Industry Building in a Smart City},
  author  = {V. E. Sathishkumar and Changsun Shin and Yongyun Cho},
  journal = {Building Research \& Information},
  volume  = {49},
  number  = {1},
  pages   = {127--143},
  year    = {2021},
  doi     = {10.1080/09613218.2020.1811908}
}

@article{wang2024deep,
  title         = {Deep Time Series Models: A Comprehensive Survey and Benchmark},
  author        = {Yuxuan Wang and Haixu Wu and Jiaxiang Dong and Yong Liu and Mingsheng Long and Jianmin Wang},
  journal       = {arXiv preprint arXiv:2407.13278},
  year          = {2024}
}

@article{lim2021forecasting,
  title   = {Time Series Forecasting With Deep Learning: A Survey},
  author  = {Bryan Lim and Stefan Zohren},
  journal = {Philosophical Transactions of the Royal Society A: Mathematical, Physical and Engineering Sciences},
  volume  = {379},
  number  = {2194},
  pages   = {20200209},
  year    = {2021},
  doi     = {10.1098/rsta.2020.0209}
}

@inproceedings{wu2021autoformer,
  title     = {Autoformer: Decomposition Transformers with Auto-Correlation for Long-Term Series Forecasting},
  author    = {Haixu Wu and Jiehui Xu and Jianmin Wang and Mingsheng Long},
  booktitle = {Advances in Neural Information Processing Systems},
  year      = {2021}
}

@inproceedings{zhou2022fedformer,
  title     = {FEDformer: Frequency Enhanced Decomposed Transformer for Long-term Series Forecasting},
  author    = {Tian Zhou and Ziqing Ma and Qingsong Wen and Xue Wang and Liang Sun and Rong Jin},
  booktitle = {Proceedings of the 39th International Conference on Machine Learning},
  pages     = {27268--27286},
  publisher = {PMLR},
  year      = {2022}
}

@inproceedings{nie2023patchtst,
  title     = {A Time Series is Worth 64 Words: Long-term Forecasting with Transformers},
  author    = {Yuqi Nie and Nam H. Nguyen and Phanwadee Sinthong and Jayant Kalagnanam},
  booktitle = {International Conference on Learning Representations},
  year      = {2023}
}

@inproceedings{liu2024itransformer,
  title     = {iTransformer: Inverted Transformers Are Effective for Time Series Forecasting},
  author    = {Yong Liu and Tengge Hu and Haoran Zhang and Haixu Wu and Shiyu Wang and Lintao Ma and Mingsheng Long},
  booktitle = {International Conference on Learning Representations},
  year      = {2024}
}

@article{he2021automl,
  title   = {{AutoML}: A Survey of the State-of-the-Art},
  author  = {Xin He and Kaiyong Zhao and Xiaowen Chu},
  journal = {Knowledge-Based Systems},
  volume  = {212},
  pages   = {106622},
  year    = {2021},
  doi     = {10.1016/j.knosys.2020.106622}
}

@article{elsken2019nas,
  title   = {Neural Architecture Search: A Survey},
  author  = {Thomas Elsken and Jan Hendrik Metzen and Frank Hutter},
  journal = {Journal of Machine Learning Research},
  volume  = {20},
  number  = {55},
  pages   = {1--21},
  year    = {2019}
}

@article{lindauer2020bestnas,
  title   = {Best Practices for Scientific Research on Neural Architecture Search},
  author  = {Marius Lindauer and Frank Hutter},
  journal = {Journal of Machine Learning Research},
  volume  = {21},
  number  = {243},
  pages   = {1--18},
  year    = {2020}
}

@inproceedings{jin2024timellm,
  title     = {Time-LLM: Time Series Forecasting by Reprogramming Large Language Models},
  author    = {Ming Jin and Shiyu Wang and Lintao Ma and Zhixuan Chu and James Y. Zhang and Xiaoming Shi and Pin-Yu Chen and Yuxuan Liang and Yuan-Fang Li and Shirui Pan and Qingsong Wen},
  booktitle = {International Conference on Learning Representations},
  year      = {2024}
}

@inproceedings{liu2025timecma,
  title     = {{TimeCMA}: Towards {LLM}-Empowered Multivariate Time Series Forecasting via Cross-Modality Alignment},
  author    = {Chenxi Liu and Qianxiong Xu and Hao Miao and Sun Yang and Lingzheng Zhang and Cheng Long and Ziyue Li and Rui Zhao},
  booktitle = {Proceedings of the AAAI Conference on Artificial Intelligence},
  volume    = {39},
  pages     = {18780--18788},
  year      = {2025},
  doi       = {10.1609/aaai.v39i18.34067}
}

@inproceedings{zhong2025timevlm,
  title     = {{Time-VLM}: Exploring Multimodal Vision-Language Models for Augmented Time Series Forecasting},
  author    = {Siru Zhong and Weilin Ruan and Ming Jin and Huan Li and Qingsong Wen and Yuxuan Liang},
  booktitle = {Proceedings of the 42nd International Conference on Machine Learning},
  year      = {2025},
  url       = {https://arxiv.org/abs/2502.04395}
}

@inproceedings{liu2025sundial,
  title     = {Sundial: A Family of Highly Capable Time Series Foundation Models},
  author    = {Yong Liu and Guo Qin and Zhiyuan Shi and Zhi Chen and Caiyin Yang and Xiangdong Huang and Jianmin Wang and Mingsheng Long},
  booktitle = {Proceedings of the 42nd International Conference on Machine Learning},
  series    = {Proceedings of Machine Learning Research},
  volume    = {267},
  pages     = {39295--39317},
  publisher = {PMLR},
  year      = {2025},
  url       = {https://proceedings.mlr.press/v267/liu25be.html}
}

@misc{datadog2026toto2,
  title         = {Toto 2.0: Time Series Forecasting Enters the Scaling Era},
  author        = {Emaad Khwaja and Chris Lettieri and Gerald Woo and Eden Belouadah and Marc Cenac and Guillaume Jarry and Enguerrand Paquin and Xunyi Zhao and Viktoriya Zhukov and Othmane Abou-Amal and Chenghao Liu and Ameet Talwalkar and David Asker},
  year          = {2026},
  eprint        = {2605.20119},
  archivePrefix = {arXiv},
  primaryClass  = {cs.LG},
  url           = {https://arxiv.org/abs/2605.20119}
}

@inproceedings{huang2024mlagentbench,
  title     = {{MLAgentBench}: Evaluating Language Agents on Machine Learning Experimentation},
  author    = {Qian Huang and Jian Vora and Percy Liang and Jure Leskovec},
  booktitle = {Proceedings of the 41st International Conference on Machine Learning},
  pages     = {20271--20309},
  publisher = {PMLR},
  year      = {2024}
}

@misc{jiang2025aide,
  title         = {AIDE: {AI}-Driven Exploration in the Space of Code},
  author        = {Zhengyao Jiang and Dominik Schmidt and Dhruv Srikanth and Dixing Xu and Ian Kaplan and Deniss Jacenko and Yuxiang Wu},
  year          = {2025},
  eprint        = {2502.13138},
  archivePrefix = {arXiv},
  primaryClass  = {cs.AI},
  url           = {https://arxiv.org/abs/2502.13138}
}

@inproceedings{yang2024sweagent,
  title     = {{SWE}-agent: Agent-Computer Interfaces Enable Automated Software Engineering},
  author    = {John Yang and Carlos E. Jimenez and Alexander Wettig and Kilian Lieret and Shunyu Yao and Karthik R. Narasimhan and Ofir Press},
  booktitle = {Advances in Neural Information Processing Systems},
  year      = {2024},
  url       = {https://arxiv.org/abs/2405.15793}
}

@misc{yang2025rdagent,
  title         = {{R\&D-Agent}: An {LLM}-Agent Framework Towards Autonomous Data Science},
  author        = {Xu Yang and Xiao Yang and Shikai Fang and Yifei Zhang and Jian Wang and Bowen Xian and Qizheng Li and Jingyuan Li and Minrui Xu and Yuante Li and Haoran Pan and Yuge Zhang and Weiqing Liu and Yelong Shen and Weizhu Chen and Jiang Bian},
  year          = {2025},
  eprint        = {2505.14738},
  archivePrefix = {arXiv},
  primaryClass  = {cs.AI},
  url           = {https://arxiv.org/abs/2505.14738}
}

@misc{kumar2026autoresearch,
  title         = {AutoResearch: An Execution-Grounded Multi-Agent Framework for Reliable Research Workflow Automation},
  author        = {Rajesh Kumar and Waqar Ali and Junaid Ahmed and Abdullah Aman Khan and Shaoning Zeng},
  year          = {2026},
  eprint        = {2607.02520},
  archivePrefix = {arXiv},
  primaryClass  = {cs.CY},
  url           = {https://arxiv.org/abs/2607.02520}
}

@misc{zhao2025timeseriesscientist,
  title         = {TimeSeriesScientist: A General-Purpose AI Agent for Time Series Analysis},
  author        = {Haokun Zhao and Xiang Zhang and Jiaqi Wei and Yiwei Xu and Yuting He and Siqi Sun and Chenyu You},
  year          = {2025},
  eprint        = {2510.01538},
  archivePrefix = {arXiv},
  primaryClass  = {cs.LG},
  url           = {https://arxiv.org/abs/2510.01538}
}

@misc{xu2026seats,
  title         = {{SEA-TS}: Self-Evolving Agent for Autonomous Code Generation of Time Series Forecasting Algorithms},
  author        = {Longkun Xu and Xiaochun Zhang and Qiantu Tuo and Rui Li},
  year          = {2026},
  eprint        = {2603.04873},
  archivePrefix = {arXiv},
  primaryClass  = {cs.AI},
  url           = {https://arxiv.org/abs/2603.04873}
}

@misc{yang2024nader,
  title         = {{NADER}: Neural Architecture Design via Multi-Agent Collaboration},
  author        = {Zekang Yang and Wang Zeng and Sheng Jin and Chen Qian and Ping Luo and Wentao Liu},
  year          = {2024},
  eprint        = {2412.19206},
  archivePrefix = {arXiv},
  primaryClass  = {cs.LG},
  url           = {https://arxiv.org/abs/2412.19206}
}

@misc{jeong2026agentnas,
  title         = {Agentic Neural Architecture Search},
  author        = {Seokhoon Jeong and Mijung Kim and Taehwan Kim},
  year          = {2026},
  eprint        = {2607.07984},
  archivePrefix = {arXiv},
  primaryClass  = {cs.LG},
  url           = {https://arxiv.org/abs/2607.07984}
}

@misc{poikarayil2026genautoml,
  title         = {{GenAutoML}: An Agentic Framework for Dynamic Architecture Generation and Optimization in Time-Series Analysis},
  author        = {Oleeviya Babu Poikarayil and C{\'e}dric Schockaert and Abdulrahman Nahhas and Christian Daase and Mursal Dawodi and Jawid Ahmad Baktash},
  year          = {2026},
  eprint        = {2606.05860},
  archivePrefix = {arXiv},
  primaryClass  = {cs.LG},
  url           = {https://arxiv.org/abs/2606.05860}
}

@misc{liu2025tsagent,
  title         = {{TS-Agent}: Understanding and Reasoning Over Raw Time Series via Iterative Insight Gathering},
  author        = {Penghang Liu and Elizabeth Fons and Annita Vapsi and Mohsen Ghassemi and Svitlana Vyetrenko and Daniel Borrajo and Vamsi K. Potluru and Manuela Veloso},
  year          = {2025},
  eprint        = {2510.07432},
  archivePrefix = {arXiv},
  primaryClass  = {cs.LG},
  url           = {https://arxiv.org/abs/2510.07432}
}

@misc{das2026nexus,
  title         = {Nexus: An Agentic Framework for Time Series Forecasting},
  author        = {Sarkar Snigdha Sarathi Das and Palash Goyal and Mihir Parmar and Nanyun Peng and Vishy Tirumalashetty and Chun-Liang Li and Rui Zhang and Jinsung Yoon and Tomas Pfister},
  year          = {2026},
  eprint        = {2605.14389},
  archivePrefix = {arXiv},
  primaryClass  = {cs.LG},
  url           = {https://arxiv.org/abs/2605.14389}
}

@misc{nam2025mlestar,
  title         = {{MLE-STAR}: Machine Learning Engineering Agent via Search and Targeted Refinement},
  author        = {Jaehyun Nam and Jinsung Yoon and Jiefeng Chen and Jinwoo Shin and Sercan {\"O}. Arik and Tomas Pfister},
  year          = {2025},
  eprint        = {2506.15692},
  archivePrefix = {arXiv},
  primaryClass  = {cs.LG},
  url           = {https://arxiv.org/abs/2506.15692}
}

@misc{zhang2025internagent,
  title         = {InternAgent: When Agent Becomes the Scientist -- Building Closed-Loop System from Hypothesis to Verification},
  author        = {{InternAgent Team} and Bo Zhang and Shiyang Feng and Xiangchao Yan and Jiakang Yuan and Runmin Ma and Yusong Hu and Zhiyin Yu and Xiaohan He and Songtao Huang and Shaowei Hou and Zheng Nie and Zhilong Wang and Jinyao Liu and Tianshuo Peng and Peng Ye and Dongzhan Zhou and Shufei Zhang and Xiaosong Wang and Yilan Zhang and Meng Li and Zhongying Tu and Xiangyu Yue and Wangli Ouyang and Bowen Zhou and Lei Bai},
  year          = {2025},
  eprint        = {2505.16938},
  archivePrefix = {arXiv},
  primaryClass  = {cs.AI},
  url           = {https://arxiv.org/abs/2505.16938}
}

@misc{lu2024aiscientist,
  title         = {The {AI} Scientist: Towards Fully Automated Open-Ended Scientific Discovery},
  author        = {Chris Lu and Cong Lu and Robert Tjarko Lange and Jakob Foerster and Jeff Clune and David Ha},
  year          = {2024},
  eprint        = {2408.06292},
  archivePrefix = {arXiv},
  primaryClass  = {cs.AI},
  url           = {https://arxiv.org/abs/2408.06292}
}

@misc{yamada2025aiscientistv2,
  title         = {The {AI} Scientist-v2: Workshop-Level Automated Scientific Discovery via Agentic Tree Search},
  author        = {Yutaro Yamada and Robert Tjarko Lange and Cong Lu and Shengran Hu and Chris Lu and Jakob Foerster and Jeff Clune and David Ha},
  year          = {2025},
  eprint        = {2504.08066},
  archivePrefix = {arXiv},
  primaryClass  = {cs.AI},
  url           = {https://arxiv.org/abs/2504.08066}
}

@misc{xia2026researchloop,
  title         = {ResearchLoop: An Evidence-Gated Control Plane for {AI}-Assisted Research},
  author        = {Yihan Xia and Taotao Wang},
  year          = {2026},
  eprint        = {2605.28282},
  archivePrefix = {arXiv},
  primaryClass  = {cs.AI},
  url           = {https://arxiv.org/abs/2605.28282}
}

@misc{hambardzumyan2026aira2,
  title         = {{AIRA}\_2: Overcoming Bottlenecks in {AI} Research Agents},
  author        = {Karen Hambardzumyan and Nicolas Baldwin and Edan Toledo and Rishi Hazra and Michael Kuchnik and Bassel Al Omari and Thomas Simon Foster and Anton Protopopov and Jean-Christophe Gagnon-Audet and Ishita Mediratta and Kelvin Niu and Michael Shvartsman and Alisia Lupidi and Alexis Audran-Reiss and Parth Pathak and Tatiana Shavrina and Despoina Magka and Hela Momand and Derek Dunfield and Nicola Cancedda and Pontus Stenetorp and Carole-Jean Wu and Jakob Nicolaus Foerster and Yoram Bachrach and Martin Josifoski},
  year          = {2026},
  eprint        = {2603.26499},
  archivePrefix = {arXiv},
  primaryClass  = {cs.AI},
  url           = {https://arxiv.org/abs/2603.26499}
}

@misc{zou2026fmlbench,
  title         = {{FML}-bench: A Controlled Study of {AI} Research Agent Strategies from the Perspective of Search Dynamics},
  author        = {Qiran Zou and Hou Hei Lam and Wenhao Zhao and Tingting Chen and Yiming Tang and Samson Yu and Yingtao Zhu and Srinivas Anumasa and Zufeng Zhang and Tianyi Zhang and Chang Liu and Zhengyao Jiang and Anirudh Goyal and Dianbo Liu},
  year          = {2026},
  eprint        = {2605.17373},
  archivePrefix = {arXiv},
  primaryClass  = {cs.LG},
  url           = {https://arxiv.org/abs/2605.17373}
}

@misc{garikaparthi2026researchgym,
  title         = {ResearchGym: Evaluating Language Model Agents on Real-World {AI} Research},
  author        = {Aniketh Garikaparthi and Manasi Patwardhan and Arman Cohan},
  year          = {2026},
  eprint        = {2602.15112},
  archivePrefix = {arXiv},
  primaryClass  = {cs.AI},
  url           = {https://arxiv.org/abs/2602.15112}
}

@misc{jin2026arbor,
  title         = {Toward Generalist Autonomous Research via Hypothesis-Tree Refinement},
  author        = {Jiajie Jin and Yuyang Hu and Kai Qiu and Qi Dai and Chong Luo and Guanting Dong and Xiaoxi Li and Tong Zhao and Xiaolong Ma and Gongrui Zhang and Zhirong Wu and Bei Liu and Zhengyuan Yang and Linjie Li and Lijuan Wang and Hongjin Qian and Yutao Zhu and Zhicheng Dou},
  year          = {2026},
  eprint        = {2606.11926},
  archivePrefix = {arXiv},
  primaryClass  = {cs.AI},
  url           = {https://arxiv.org/abs/2606.11926}
}

@inproceedings{zeng2023are,
  title     = {Are Transformers Effective for Time Series Forecasting?},
  author    = {Ailing Zeng and Muxi Chen and Lei Zhang and Qiang Xu},
  booktitle = {Proceedings of the AAAI Conference on Artificial Intelligence},
  volume    = {37},
  pages     = {11121--11128},
  year      = {2023}
}

@inproceedings{si2025cmos,
  title     = {{CMoS}: Rethinking Time Series Prediction Through the Lens of Chunk-wise Spatial Correlations},
  author    = {Haotian Si and Changhua Pei and Jianhui Li and Dan Pei and Gaogang Xie},
  booktitle = {Proceedings of the 42nd International Conference on Machine Learning},
  series    = {Proceedings of Machine Learning Research},
  volume    = {267},
  pages     = {55493--55510},
  publisher = {PMLR},
  url       = {https://proceedings.mlr.press/v267/si25a.html},
  year      = {2025}
}

@inproceedings{zhou2025crosslinear,
  title     = {CrossLinear: Plug-and-Play Cross-Correlation Embedding for Time Series Forecasting with Exogenous Variables},
  author    = {Pengfei Zhou and Yunlong Liu and Junli Liang and Qi Song and Xiangyang Li},
  booktitle = {Proceedings of the 31st ACM SIGKDD Conference on Knowledge Discovery and Data Mining},
  year      = {2025},
  doi       = {10.1145/3711896.3736899}
}

@inproceedings{lin2024sparsetsf,
  title     = {SparseTSF: Modeling Long-term Time Series Forecasting with 1k Parameters},
  author    = {Shengsheng Lin and Weiwei Lin and Wentai Wu and Haojun Chen and Junjie Yang},
  booktitle = {Proceedings of the 41st International Conference on Machine Learning},
  year      = {2024}
}

@inproceedings{huang2025timebase,
  title     = {TimeBase: The Power of Minimalism in Efficient Long-term Time Series Forecasting},
  author    = {Qihe Huang and Zhengyang Zhou and Kuo Yang and Zhongchao Yi and Xu Wang and Yang Wang},
  booktitle = {Proceedings of the 42nd International Conference on Machine Learning},
  series    = {Proceedings of Machine Learning Research},
  volume    = {267},
  pages     = {26227--26246},
  publisher = {PMLR},
  url       = {https://proceedings.mlr.press/v267/huang25az.html},
  year      = {2025}
}

@inproceedings{zhang2023crossformer,
  title     = {Crossformer: Transformer Utilizing Cross-Dimension Dependency for Multivariate Time Series Forecasting},
  author    = {Yunhao Zhang and Junchi Yan},
  booktitle = {International Conference on Learning Representations},
  year      = {2023}
}

@inproceedings{lu2026dtaf,
  title     = {Towards Non-Stationary Time Series Forecasting with Temporal Stabilization and Frequency Differencing},
  author    = {Junkai Lu and Peng Chen and Chenjuan Guo and Yang Shu and Meng Wang and Bin Yang},
  booktitle = {Proceedings of the AAAI Conference on Artificial Intelligence},
  volume    = {40},
  pages     = {24070--24078},
  doi       = {10.1609/aaai.v40i29.39585},
  year      = {2026}
}

@inproceedings{qiu2025duet,
  title     = {DUET: Dual Clustering Enhanced Multivariate Time Series Forecasting},
  author    = {Xiangfei Qiu and Xingjian Wu and Yan Lin and Chenjuan Guo and Jilin Hu and Bin Yang},
  booktitle = {Proceedings of the 31st ACM SIGKDD Conference on Knowledge Discovery and Data Mining},
  pages     = {1185--1196},
  doi       = {10.1145/3690624.3709325},
  year      = {2025}
}

@inproceedings{woo2022etsformer,
  title     = {ETSformer: Exponential Smoothing Transformers for Time-series Forecasting},
  author    = {Gerald Woo and Chenghao Liu and Doyen Sahoo and Akshat Kumar and Steven C. H. Hoi},
  booktitle = {International Conference on Learning Representations},
  year      = {2022}
}

@inproceedings{zhou2021informer,
  title     = {Informer: Beyond Efficient Transformer for Long Sequence Time-Series Forecasting},
  author    = {Haoyi Zhou and Shanghang Zhang and Jieqi Peng and Shuai Zhang and Jianxin Li and Hui Xiong and Wancai Zhang},
  booktitle = {Proceedings of the AAAI Conference on Artificial Intelligence},
  volume    = {35},
  pages     = {11106--11115},
  year      = {2021}
}

@article{naghashi2025multipatchformer,
  title   = {A Multiscale Model for Multivariate Time Series Forecasting},
  author  = {Vahid Naghashi and Mounir Boukadoum and Abdoulaye Banire Diallo},
  journal = {Scientific Reports},
  volume  = {15},
  pages   = {1565},
  year    = {2025},
  doi     = {10.1038/s41598-024-82417-4}
}

@inproceedings{liu2022nonstationary,
  title     = {Non-stationary Transformers: Exploring the Stationarity in Time Series Forecasting},
  author    = {Yong Liu and Haixu Wu and Jianmin Wang and Mingsheng Long},
  booktitle = {Advances in Neural Information Processing Systems},
  year      = {2022}
}

@inproceedings{tan2024arellms,
  title     = {Are Language Models Actually Useful for Time Series Forecasting?},
  author    = {Mingtian Tan and Mike A. Merrill and Vinayak Gupta and Tim Althoff and Thomas Hartvigsen},
  booktitle = {Advances in Neural Information Processing Systems},
  year      = {2024}
}

@inproceedings{dai2024pdf,
  title     = {Periodicity Decoupling Framework for Long-term Series Forecasting},
  author    = {Tao Dai and Beiliang Wu and Peiyuan Liu and Naiqi Li and Jigang Bao and Yu Jiang and Shu-Tao Xia},
  booktitle = {International Conference on Learning Representations},
  year      = {2024}
}

@inproceedings{chen2024pathformer,
  title     = {Pathformer: Multi-scale Transformers with Adaptive Pathways for Time Series Forecasting},
  author    = {Peng Chen and Yingying Zhang and Yunyao Cheng and Yang Shu and Yihang Wang and Qingsong Wen and Bin Yang and Chenjuan Guo},
  booktitle = {International Conference on Learning Representations},
  year      = {2024}
}

@inproceedings{liu2022pyraformer,
  title     = {Pyraformer: Low-complexity Pyramidal Attention for Long-range Time Series Modeling and Forecasting},
  author    = {Shizhan Liu and Hang Yu and Cong Liao and Jianmin Wang and Weihua Lin and Alex X. Liu},
  booktitle = {International Conference on Learning Representations},
  year      = {2022}
}

@article{lim2021tft,
  title   = {Temporal Fusion Transformers for Interpretable Multi-horizon Time Series Forecasting},
  author  = {Bryan Lim and Sercan O. Arik and Nicolas Loeff and Tomas Pfister},
  journal = {International Journal of Forecasting},
  volume  = {37},
  number  = {4},
  pages   = {1748--1764},
  year    = {2021}
}

@inproceedings{liu2025timebridge,
  title     = {TimeBridge: Non-Stationarity Matters for Long-term Time Series Forecasting},
  author    = {Peiyuan Liu and Beiliang Wu and Yifan Hu and Naiqi Li and Tao Dai and Jigang Bao and Shu-Tao Xia},
  booktitle = {Proceedings of the 42nd International Conference on Machine Learning},
  year      = {2025}
}

@inproceedings{lee2025timeperceiver,
  title     = {TimePerceiver: An Encoder-Decoder Framework for Generalized Time-Series Forecasting},
  author    = {Jaebin Lee and Hankook Lee},
  booktitle = {Advances in Neural Information Processing Systems},
  year      = {2025},
  url       = {https://proceedings.neurips.cc/paper_files/paper/2025/hash/c6c682ba9bd8839104f2a82901da4109-Abstract-Conference.html}
}

@inproceedings{vaswani2017attention,
  title     = {Attention Is All You Need},
  author    = {Ashish Vaswani and Noam Shazeer and Niki Parmar and Jakob Uszkoreit and Llion Jones and Aidan N. Gomez and Lukasz Kaiser and Illia Polosukhin},
  booktitle = {Advances in Neural Information Processing Systems},
  year      = {2017}
}

@inproceedings{cirstea2022triformer,
  title     = {Triformer: Triangular, Variable-Specific Attentions for Long Sequence Multivariate Time Series Forecasting},
  author    = {Razvan-Gabriel Cirstea and Chenjuan Guo and Bin Yang and Tung Kieu and Xuanyi Dong and Shirui Pan},
  booktitle = {Proceedings of the Thirty-First International Joint Conference on Artificial Intelligence},
  pages     = {1994--2001},
  year      = {2022}
}

@inproceedings{wang2024timemixer,
  title     = {TimeMixer: Decomposable Multiscale Mixing for Time Series Forecasting},
  author    = {Shiyu Wang and Haixu Wu and Xiaoming Shi and Tengge Hu and Huakun Luo and Lintao Ma and James Y. Zhang and Jun Zhou},
  booktitle = {International Conference on Learning Representations},
  year      = {2024}
}

@inproceedings{murad2025wpmixer,
  title     = {WPMixer: Efficient Multi-Resolution Mixing for Long-Term Time Series Forecasting},
  author    = {Md Mahmuddun Nabi Murad and Mehmet Aktukmak and Yasin Yilmaz},
  booktitle = {Proceedings of the AAAI Conference on Artificial Intelligence},
  volume    = {39},
  pages     = {19581--19588},
  year      = {2025}
}

@inproceedings{wang2023micn,
  title     = {MICN: Multi-scale Local and Global Context Modeling for Long-term Series Forecasting},
  author    = {Huiqiang Wang and Jian Peng and Feihu Huang and Jince Wang and Junhui Chen and Yifei Xiao},
  booktitle = {International Conference on Learning Representations},
  year      = {2023}
}

@inproceedings{luo2024moderntcn,
  title     = {ModernTCN: A Modern Pure Convolution Structure for General Time Series Analysis},
  author    = {Donghao Luo and Xue Wang},
  booktitle = {International Conference on Learning Representations},
  year      = {2024}
}

@inproceedings{wu2023timesnet,
  title     = {TimesNet: Temporal 2D-Variation Modeling for General Time Series Analysis},
  author    = {Haixu Wu and Tengge Hu and Yong Liu and Hang Zhou and Jianmin Wang and Mingsheng Long},
  booktitle = {International Conference on Learning Representations},
  year      = {2023}
}

@inproceedings{zhang2024msgnet,
  title     = {MSGNet: Learning Multi-Scale Inter-Series Correlations for Multivariate Time Series Forecasting},
  author    = {Qingsong Wen and Tian Zhou and Chaoli Zhang and Weixin Xie and Liang Sun and Ziqing Ma and James Y. Zhang and Rong Jin},
  booktitle = {Proceedings of the AAAI Conference on Artificial Intelligence},
  volume    = {38},
  pages     = {11010--11018},
  year      = {2024}
}

@inproceedings{kitaev2020reformer,
  title     = {Reformer: The Efficient Transformer},
  author    = {Nikita Kitaev and {\L}ukasz Kaiser and Anselm Levskaya},
  booktitle = {International Conference on Learning Representations},
  year      = {2020}
}

@inproceedings{yi2023frets,
  title     = {Frequency-domain {MLP}s Are More Effective Learners in Time Series Forecasting},
  author    = {Kun Yi and Qi Zhang and Wei Fan and Shoujin Wang and Pengyang Wang and Hui He and Defu Lian and Ning An and Longbing Cao and Zhendong Niu},
  booktitle = {Advances in Neural Information Processing Systems},
  year      = {2023}
}

@inproceedings{huang2024hdmixer,
  title     = {{HDMixer}: Hierarchical Dependency with Extendable Patch for Multivariate Time Series Forecasting},
  author    = {Qihe Huang and Lei Shen and Ruixin Zhang and Jiahuan Cheng and Shouhong Ding and Zhengyang Zhou and Yang Wang},
  booktitle = {Proceedings of the AAAI Conference on Artificial Intelligence},
  volume    = {38},
  pages     = {12608--12616},
  doi       = {10.1609/aaai.v38i11.29155},
  year      = {2024}
}

@inproceedings{tang2025patchmlp,
  title     = {Unlocking the Power of Patch: Patch-Based {MLP} for Long-Term Time Series Forecasting},
  author    = {Peiwang Tang and Weitai Zhang},
  booktitle = {Proceedings of the AAAI Conference on Artificial Intelligence},
  volume    = {39},
  pages     = {12640--12648},
  doi       = {10.1609/aaai.v39i12.33378},
  year      = {2025}
}

@inproceedings{hu2025timefilter,
  title     = {TimeFilter: Patch-Specific Spatial-Temporal Graph Filtration for Time Series Forecasting},
  author    = {Yifan Hu and Guibin Zhang and Peiyuan Liu and Disen Lan and Naiqi Li and Dawei Cheng and Tao Dai and Shu-Tao Xia and Shirui Pan},
  booktitle = {Proceedings of the 42nd International Conference on Machine Learning},
  series    = {Proceedings of Machine Learning Research},
  volume    = {267},
  pages     = {24893--24911},
  publisher = {PMLR},
  url       = {https://proceedings.mlr.press/v267/hu25ac.html},
  year      = {2025}
}

@inproceedings{fei2025amplifier,
  title     = {Amplifier: Bringing Attention to Neglected Low-Energy Components in Time Series Forecasting},
  author    = {Jingru Fei and Kun Yi and Wei Fan and Qi Zhang and Zhendong Niu},
  booktitle = {Proceedings of the AAAI Conference on Artificial Intelligence},
  volume    = {39},
  pages     = {11645--11653},
  doi       = {10.1609/aaai.v39i11.33267},
  year      = {2025}
}

@inproceedings{xu2024fits,
  title     = {FITS: Modeling Time Series with 10k Parameters},
  author    = {Zhijian Xu and Ailing Zeng and Qiang Xu},
  booktitle = {International Conference on Learning Representations},
  year      = {2024}
}

@inproceedings{zhou2022film,
  title     = {FiLM: Frequency Improved Legendre Memory Model for Long-term Time Series Forecasting},
  author    = {Tian Zhou and Ziqing Ma and Xue Wang and Qingsong Wen and Liang Sun and Tao Yao and Wotao Yin and Rong Jin},
  booktitle = {Advances in Neural Information Processing Systems},
  year      = {2022}
}

@inproceedings{zhou2025kanad,
  title     = {{KAN-AD}: Time Series Anomaly Detection with Kolmogorov-Arnold Networks},
  author    = {Quan Zhou and Changhua Pei and Fei Sun and Han Jing and Zhengwei Gao and Haiming Zhang and Gaogang Xie and Dan Pei and Jianhui Li},
  booktitle = {Proceedings of the 42nd International Conference on Machine Learning},
  series    = {Proceedings of Machine Learning Research},
  volume    = {267},
  pages     = {79136--79149},
  publisher = {PMLR},
  url       = {https://proceedings.mlr.press/v267/zhou25u.html},
  year      = {2025}
}

@article{liu2023koopa,
  title   = {Koopa: Learning Non-stationary Time Series Dynamics with Koopman Predictors},
  author  = {Yong Liu and Chenyu Li and Jianmin Wang and Mingsheng Long},
  journal = {arXiv preprint arXiv:2305.18803},
  year    = {2023}
}

@article{gu2023mamba,
  title   = {Mamba: Linear-Time Sequence Modeling with Selective State Spaces},
  author  = {Albert Gu and Tri Dao},
  journal = {arXiv preprint arXiv:2312.00752},
  year    = {2023}
}

@inproceedings{wu2025srsnet,
  title     = {Enhancing Time Series Forecasting through Selective Representation Spaces: A Patch Perspective},
  author    = {Xingjian Wu and Xiangfei Qiu and Hanyin Cheng and Zhengyu Li and Jilin Hu and Chenjuan Guo and Bin Yang},
  booktitle = {Advances in Neural Information Processing Systems},
  year      = {2025},
  url       = {https://proceedings.neurips.cc/paper_files/paper/2025/hash/21beeac772f6f84d1517a2fc81e582d0-Abstract-Conference.html}
}

@article{lin2026segrnn,
  title   = {SegRNN: Segment Recurrent Neural Network for Long-Term Time Series Forecasting},
  author  = {Shengsheng Lin and Weiwei Lin and Wentai Wu and Feiyu Zhao and Ruichao Mo and Haotong Zhang},
  journal = {IEEE Internet of Things Journal},
  volume  = {13},
  number  = {5},
  pages   = {9861--9871},
  year    = {2026},
  doi     = {10.1109/JIOT.2025.3647705}
}

@article{huang2025timekan,
  title   = {TimeKAN: KAN-based Frequency Decomposition Learning Architecture for Long-term Time Series Forecasting},
  author  = {Songtao Huang and Zhen Zhao and Can Li and Lei Bai},
  journal = {arXiv preprint arXiv:2502.06910},
  year    = {2025}
}

@inproceedings{stitsyuk2025xpatch,
  title     = {xPatch: Dual-Stream Time Series Forecasting with Exponential Seasonal-Trend Decomposition},
  author    = {Artyom Stitsyuk and Jaesik Choi},
  booktitle = {Proceedings of the AAAI Conference on Artificial Intelligence},
  year      = {2025}
}

@article{zhang2022lightts,
  title   = {Less Is More: Fast Multivariate Time Series Forecasting with Light Sampling-oriented MLP Structures},
  author  = {Xingjian Zhang and Xuanze Zhang and Shiyang Zhang and Tian Zhou and Yiqun Zhang and Furu Wei and Yanfeng Wang},
  journal = {arXiv preprint arXiv:2207.01186},
  year    = {2022}
}

@article{chen2023tsmixer,
  title   = {TSMixer: An All-MLP Architecture for Time Series Forecasting},
  author  = {Si-An Chen and Chun-Liang Li and Nate Yoder and Sercan O. Arik and Tomas Pfister},
  journal = {Transactions on Machine Learning Research},
  year    = {2023}
}

@article{das2023tide,
  title   = {Long-term Forecasting with TiDE: Time-series Dense Encoder},
  author  = {Abhimanyu Das and Weihao Kong and Rajat Sen and Yichen Zhou},
  journal = {Transactions on Machine Learning Research},
  year    = {2023}
}

@article{liu2022scinet,
  title   = {SCINet: Time Series Modeling and Forecasting with Sample Convolution and Interaction},
  author  = {Minhao Liu and Ailing Zeng and Zhijian Xu and Qiuxia Lai and Qiang Xu and Jiaya Jia},
  journal = {Advances in Neural Information Processing Systems},
  year    = {2022}
}
